%% file: bare_jrnl_new_sample4.tex
\documentclass[lettersize,journal]{IEEEtran}
\usepackage{amsmath,amsfonts}
\usepackage{algorithmic}
\usepackage{algorithm}
\usepackage{array}
\usepackage[caption=false,font=normalsize,labelfont=sf,textfont=sf]{subfig}
\usepackage{textcomp}
\usepackage{stfloats}
\usepackage{url}
\usepackage{verbatim}
\usepackage{graphicx}
\usepackage{cite}
\usepackage{tikz}
\usetikzlibrary{arrows.meta,positioning}
\usetikzlibrary{arrows.meta, positioning, shadows}
\usepackage{graphicx}
\usepackage[table]{xcolor}
\usepackage{tabularx}
\usepackage{colortbl}
\usepackage{booktabs}
\usepackage[colorlinks=true, citecolor=blue, linkcolor=black, hypertexnames=false]{hyperref}
\begin{document}

\title{Exteroceptive Sensor Faults in Autonomous Vehicles: 
A Cross-Modal Taxonomy and Fault Injection Toolkit}

\author{Davide Possenti$^{1,*}$, Daniele Cianca$^{1}$, Stefano Arrigoni$^{1}$, Francesco Braghin$^{1}$
        % <-this % stops a space
\thanks{$^{1}$Davide Possenti, Daniele Cianca, Stefano Arrigoni and Francesco Braghin are with Department of Mechanical Engineering, Politecnico di Milano, 20156 Milano, Italia.}%
\thanks{$^{*}$Corresponding author: Davide Possenti (e-mail: davide.possenti@polimi.it).}%

\thanks{This work has been submitted to the IEEE for possible publication. Copyright may be transferred without notice, after which this version may no longer be accessible.}}% <-this % stops a space

% The paper headers
%\markboth{Journal of \LaTeX\ Class Files,~Vol.~14, No.~8, August~2021}%
%{Shell \MakeLowercase{\textit{et al.}}: A Sample Article Using IEEEtran.cls for IEEE Journals}

\markboth{Preprint submitted to IEEE, 2026}{Author \MakeLowercase{\textit{et al.}}: Title}

%\IEEEpubid{0000--0000/00\$00.00~\copyright~2021 IEEE}
% Remember, if you use this you must call \IEEEpubidadjcol in the second
% column for its text to clear the IEEEpubid mark.

\maketitle

\begin{abstract}
Autonomous vehicles rely on cameras, LiDAR, and radar to perceive their surroundings, yet these exteroceptive sensors can be degraded by hardware aging, environmental interference, miscalibration, embedded software errors, and combinations of these causes, with direct consequences for perception quality and safety. Existing surveys and benchmarks mostly address individual sensors, isolated conditions, or vehicle-level consequences, and no common vocabulary relates fault origins across the three modalities. This paper reviews the literature on exteroceptive sensor faults and proposes a cross-modal, origin-driven taxonomy that organizes them into five categories, namely hardware, environmental, calibration, software, and synergistic faults, mapping each entry to its physical or logical mechanism, observable manifestation, and affected subsystem, and relating the categories to the safety analyses of ISO 26262 and ISO 21448. To make the taxonomy operational, we release an open-source fault injection toolkit with 94 symptom-level routines (41 camera, 27 LiDAR, and 26 radar), four graded severity levels, reproducible burst scheduling, and per-frame labels recording the fault category, affected subsystem, and severity. Examples on the Boreas dataset illustrate how each category alters camera images, LiDAR point clouds, and radar scans. 
Together, the taxonomy and the toolkit provide a shared reference and a controlled testbed for studying sensor reliability, developing and benchmarking fault detection and diagnosis methods, and building fault-annotated datasets for the safety validation of perception systems. The toolkit is publicly available at \url{https://github.com/davidepossenti/AV-SensorFault-Toolkit}.
\end{abstract}


\section{Introduction}
\label{sec:introduction}
\IEEEPARstart{P}{erception} in autonomous vehicles (AVs) relies on the complementary use of camera, radar, and LiDAR sensors to ensure robust and redundant environmental awareness. These \textit{exteroceptive sensors} serve as the primary source of environmental understanding for downstream tasks such as localization, mapping, and object detection~\cite{Yeong2021}. Despite significant advances in deep learning-based perception, the reliability of these sensors remains a major bottleneck for safe and certifiable autonomous driving. Under real-world operating conditions, sensors are susceptible to various faults during vehicular operation~\cite{FADE2025} that can distort, reduce, or entirely corrupt perception outputs.

In accordance with ISO~26262-1:2018~\cite{ISO26262_1}, a \emph{fault} is an abnormal condition that can cause an element or an item to fail. A fault may manifest itself as an \emph{error}, i.e., a discrepancy between a computed, observed, or measured value or condition and the correct one, which may in turn lead to a \emph{failure}, i.e., the termination of an intended behavior of an element or an item due to a fault manifestation. In sensing, these faults arise from different physical and logical origins: hardware degradation (e.g., optical lens damage~\cite{RGB_CameraFailures}, CMOS image sensor pixel damage~\cite{OnlineImageFaults}, or LiDAR laser diode power loss~\cite{FADE2025}), environmental interference (e.g., optical scattering from rain, snow, and fog~\cite{sensorsFog,LiDAR_snow,Dong2023}, or electromagnetic multipath reflections~\cite{Radar_Artifacts,Radar_anomaly}), miscalibration~\cite{Peng2024Miscalibration,Tahiraj2025CalorNoCal,Xia2025Misalignment}, software or image signal-processing errors~\cite{RGB_CameraFailures,Matos2025SimulatingSensorFailures}, or temporal synchronization mismatches between sensors~\cite{Dong2023}. Following common practice in the perception literature, we use the term fault broadly, also for external conditions that corrupt the sensor output without being faults of the sensor in the strict ISO sense. Such faults can lead to incomplete or misleading environmental representations, ultimately resulting in unsafe control decisions. As ISO~26262 addresses hazards caused by the malfunctioning behavior of safety-related electrical and electronic (E/E) systems, and ISO~21448 (SOTIF)~\cite{ISO21448_2022} addresses hazards arising from functional insufficiencies, including performance insufficiencies of E/E elements, a structured understanding of sensor faults and degradations has become essential.

Prior studies have explored specific fault scenarios; for instance, radar degradation under adverse weather or dust~\cite{Dust_Radar}, multipath reflections~\cite{Radar_Artifacts,Radar_anomaly}, camera anomalies~\cite{RGB_CameraFailures,FastAnomalyImage}, and LiDAR failures or environment-induced degradations~\cite{Chang2023,kong2023robo3d}. Similarly, fault diagnosis and robustness benchmarking in autonomous driving perception have gained growing attention~\cite{Dong2023,Baseline,FADE2025,SynthNoise2025}. However, existing experimental studies and injection tools typically address individual sensing modalities, at most camera and LiDAR pairs, or the driving stack as a whole rather than the sensor faults themselves~\cite{jha2018avfi,jha2019ml,maleki2023carfase}, while recent literature surveys~\cite{matos2024survey,viktor2026modality} organize anomalies primarily by individual sensor or high-level safety outcome without providing fault injection tools. Furthermore, publicly available datasets addressing adverse driving conditions focus predominantly on weather-related factors~\cite{BoreasDataset,TWICEdataset,SIDdataset,DAWNdataset,WADSdataset}, leaving internal hardware, calibration, firmware, and compound sensor faults largely underrepresented. Consequently, the community still lacks an origin-driven, cross-modal taxonomy paired with an operational toolkit to systematically inject faults and benchmark perception and diagnosis algorithms across camera, LiDAR, and radar.

To address this gap, this paper reviews the literature on exteroceptive sensor degradations and presents a \textit{cross-modal taxonomy and an open-source fault injection toolkit for exteroceptive sensor faults in autonomous vehicles}. The main contributions are threefold:

\begin{enumerate}
    \item A critical review of existing sensor fault studies and a unified, origin-driven taxonomy that categorizes faults into five classes (hardware, environmental, calibration, software, and synergistic), mapping each anomaly to its physical or logical mechanism, observable symptom, and affected subsystem across camera, LiDAR, and radar sensors;
    \item An open-source fault injection toolkit (94 routines), available at \url{https://github.com/davidepossenti/AV-SensorFault-Toolkit}, implementing the taxonomy categories across all three sensing modalities with graded severity levels, reproducible burst scheduling (a burst being a contiguous interval of faulty frames), and frame-level ground-truth annotations, to facilitate reproducible benchmarking and the creation of fault-annotated datasets;
    \item A set of qualitative visual examples, generated by injecting synthetic anomalies into time-stamped camera images, LiDAR point clouds, and radar scans of the Boreas dataset~\cite{BoreasDataset}, accompanied by a discussion of the implications for Fault Detection and Diagnosis (FDD) and functional safety analyses.
\end{enumerate}

Rather than focusing on the experimental characterization of specific hardware breakdowns, this work provides a structured conceptual framework and software resource that bridge perception engineering and safety analysis, offering a common reference for both academic research and testing of autonomous systems. 

The remainder of this paper is organized as follows. Section~\ref{sec:related_work} reviews related work on exteroceptive sensor architectures, modality-specific fault studies, fault-tolerant perception, and existing surveys. Section~\ref{sec:taxonomy} introduces the proposed cross-modal taxonomy of exteroceptive sensor faults. Section~\ref{sec:simulations} describes the open-source fault injection toolkit, presents examples of synthetically degraded sensor data on the Boreas dataset, and discusses the operational scope, design choices, and limitations of the injection routines. Section~\ref{sec:discussion} discusses the implications of the taxonomy for functional safety and fault diagnosis. Finally, Section~\ref{sec:conclusion} concludes the paper and outlines directions for future work.

\section{Related Work}
\label{sec:related_work}

Research on sensor faults in autonomous vehicles has received comparatively limited attention, despite its crucial role in environmental perception and overall safety. Existing studies predominantly address faults within individual sensing modalities or focus strictly on downstream algorithmic compensation, leaving the physical and logical origins of sensor degradation underexplored. This section reviews the fundamental architecture of exteroceptive sensors, analyzes existing studies on modality-specific faults, discusses current approaches in fault-tolerant perception and diagnosis, and positions this work with respect to existing surveys and fault injection tools.

\subsection{Sensor Architectures and Vulnerabilities}
\label{sec:sensor_architectures}

The perception systems of autonomous vehicles rely on a heterogeneous set of exteroceptive sensors to acquire data from the external world. The three primary sensing technologies, camera, LiDAR, and radar, are summarized in Table~\ref{tab:extero_sensors_taxonomy}. A structural understanding of these sensors is essential, as their physical and computational architectures dictate their specific fault vulnerabilities.

\textbf{Cameras} represent the most information-rich sensors for semantic understanding~\cite{Yeong2021}. The image-formation pipeline relies on five core subsystems: the lens, the camera body, the Bayer filter, the CMOS/CCD image sensor, and the Image Signal Processor (ISP)~\cite{CameraComponents}. Faults typically originate from optical degradation (e.g., lens damage or mechanical failure~\cite{RGB_CameraFailures}), hardware malfunctions in the pixel array (e.g., dead pixels~\cite{OnlineImageFaults}), or algorithmic failures within the ISP~\cite{RGB_CameraFailures}, which performs, among others, demosaicing and HDR processing~\cite{ISP_meets_DeepLearning,DynamicISP}. Furthermore, cameras remain intrinsically sensitive to environmental perturbations such as glare, adverse weather, and sudden illumination changes.

\textbf{LiDAR} systems provide accurate 3D spatial information by measuring the time-of-flight (ToF) or frequency shift (FMCW) of emitted laser pulses~\cite{LidarGenerico}. A LiDAR relies on the coordinated operation of laser emitters, optical lenses, photodetectors (e.g., APD or SPAD), scanning mechanisms (e.g., mechanical rotors, MEMS, or OPA), and control electronics~\cite{Giuffrida2023}. Vulnerabilities in these systems encompass emitter power loss, mechanical wear in scanning units, thermal degradation, and signal attenuation or scattering caused by fog, rain, or absorptive surfaces~\cite{Chang2023,Rob_LidarBased,Yeong2021}.

\textbf{Radar} sensors estimate the range, velocity, and angle of objects by transmitting millimeter-wave radio signals (typically 77--79~GHz, FMCW or PMCW)~\cite{Rao2025}. Unlike optical sensors, radars are largely immune to poor illumination and exhibit high resilience to adverse weather~\cite{Yeong2021}. However, radar architectures, comprising antennas, transceivers, and digital signal processors, are susceptible to hardware degradation (e.g., low-noise amplifier gain loss~\cite{RadarGracefulDegradation}), boresight misalignment, and environmental electromagnetic artifacts such as multipath reflections, clutter, and mutual interference~\cite{Srivastav2023,Giuffrida2023}.

\begin{table*}[!t]
\setlength{\tabcolsep}{3.5pt}
\caption{Exteroceptive sensors commonly used in autonomous vehicles.}
\label{tab:extero_sensors_taxonomy}
\centering
\small
\renewcommand{\arraystretch}{1.3}
\begin{tabular}{p{0.13\textwidth} p{0.22\textwidth} p{0.25\textwidth} p{0.16\textwidth} p{0.16\textwidth}}
\hline\hline
\textbf{Sensor} & \textbf{Operating Principle} & \textbf{Role in Perception} & \textbf{Advantages} & \textbf{Limitations} \\
\hline
RGB Camera~\cite{Yeong2021,matos2024survey,AdverseWeatherSurvey} & Passive optical sensing (CMOS or CCD detectors); provides 2D intensity and color images. & Semantic perception (object detection and classification, segmentation, 2D/3D tracking). & High resolution and semantic richness; low cost and weight. & Highly sensitive to lighting and weather conditions; depth available only through geometric estimation. \\[8pt]
3D LiDAR~\cite{Yeong2021,matos2024survey,AdverseWeatherSurvey,LidarGenerico} & Laser time-of-flight (or FMCW) measurement; generates 3D point clouds with range and reflectance. & Geometric perception (accurate 3D scene reconstruction, obstacle detection, SLAM). & Direct metric distance measurement; illumination independent. & High cost and size; degraded performance in fog, rain, and on dark surfaces. \\[8pt]
Radar~\cite{Yeong2021,matos2024survey,AdverseWeatherSurvey} & Millimeter-wave reflection (FMCW or PMCW); estimates range, radial velocity (Doppler), and angle. & Tracking and safety (long-range detection, relative velocity estimation, weather reliable). & Robust to weather and lighting; direct velocity measurement. & Lower spatial resolution than LiDAR; multipath effects, ghost targets, and interference. \\
\hline\hline
\end{tabular}
\end{table*}

\subsection{Modality-Specific Sensor Fault Studies}
\label{sec:review_sensor_faults}

Prior investigations into sensor reliability have predominantly isolated specific fault categories within individual sensing modalities. Cameras have received the most attention due to their critical role in scene understanding. Secci and Ceccarelli~\cite{RGB_CameraFailures} conducted a comprehensive Failure Mode and Effects Analysis (FMEA) to identify faults affecting camera subsystems. Tian et al.~\cite{FADE2025} leveraged this taxonomy to develop FADE, a fault-tolerance testing framework for autonomous driving. Other studies address specific camera degradations, such as the detection of dead pixels~\cite{OnlineImageFaults}, the synthesis and identification of noise levels in camera-radar datasets~\cite{SynthNoise2025}, and the recovery from motion blur~\cite{MotionBlurRecovery}. However, the majority of vision-based studies remain heavily skewed toward environmental conditions, such as adverse weather and raindrop occlusions~\cite{AdverseWeatherSurvey,STARNet,Robust_localization_weather,OdometryRain}.

Similarly, research on LiDAR failures has largely focused on environmental corruptions, from benchmarks such as Robo3D~\cite{kong2023robo3d} and studies on pose estimation under point cloud corruptions~\cite{Rob_LidarBased} to studies on fog~\cite{LiDAR_fog}, snow denoising~\cite{LiDAR_snow}, airborne dust~\cite{LiDAR_Dust}, crosstalk from other LiDARs~\cite{LiDAR_Crosstalk}, and frozen and wet road surfaces~\cite{LiDAR_Wet}. Chang et al.~\cite{Chang2023} broadened this scope by exploring internal hardware failures, including emitter power loss and scanning defects. Kettelgerdes et al.~\cite{Kettelgerdes2026} investigated temperature induced calibration drift in automotive LiDAR sensors. Jin et al.~\cite{Baseline} examined LiDAR-specific degradations such as field-of-view loss and crosstalk, while Dong et al.~\cite{Dong2023} benchmarked 3D object detectors against 27 synthetic corruptions spanning weather, sensor noise, and alignment errors.

In contrast, research on radar faults is comparatively limited. Ridder and Narayanan~\cite{RadarGracefulDegradation} modeled hardware degradation, such as amplifier gain reduction and phased-array element failures, for an airborne radar. Most subsequent studies focus on environmental and calibration issues: surface contamination from water and dust~\cite{Dust_Radar,Radar_Sediments}, multipath reflections causing ghost targets~\cite{Radar_Artifacts,Radar_anomaly}, and geometric boresight misalignments~\cite{RadarMisalignment}. Oh et al.~\cite{Thermal_man_Radar} investigated thermal management challenges in automotive radars, while Albrecht et al.~\cite{Albrecht2022} studied thermomechanical deformation in radar packaging. Harter et al.~\cite{HarterRadar} further addressed temperature induced calibration drift in automotive radar systems. On the processing side, studies address interference mitigation~\cite{Radar_software} and adaptive embedded processing~\cite{Radar_software2}; they identify the DSP stages whose failure would degrade radar perception, but do not study such failures.

\subsection{Robust Perception and Fault Diagnosis}
\label{sec:fusion_and_diagnosis}

To mitigate the impact of sensor degradation, recent perception frameworks exploit sensor redundancy. For 3D object detection, LiDAR-camera fusion architectures are designed to remain robust to degraded inputs~\cite{TransFusion,BEVFusion} and to sensor failures~\cite{MultiModalExpert}, and deep multimodal fusion improves detection in fog~\cite{sensorsFog}. For localization and motion estimation, selective fusion~\cite{chen2019selective,chen2022learning, talebi2025raci}, time-aware late fusion~\cite{TEFuNet}, adaptive anti-degradation~\cite{A2DO}, and graph-based~\cite{FaultResistantOdometry} and redundant~\cite{RedundantOdometry} approaches dynamically downweight or replace degraded features and estimates, ensuring robust operation under fog, rain, or partial occlusions.

While these fusion strategies successfully mask sensor degradation, they do not inherently identify the faulty sensor or characterize the fault mechanism. This limitation necessitates dedicated Fault Detection and Diagnosis (FDD) systems. Jin et al.~\cite{Baseline} proposed a hybrid framework leveraging analytical redundancy and information complexity metrics to classify predefined faults across camera and LiDAR data. Feature-level deep learning frameworks have also been developed for fault diagnosis and tolerance in multi-sensor fusion~\cite{Sensor_fault_diagnosis}, together with fault detection, identification, isolation, and health forecasting~\cite{MultiSensorFaultDetection}. Furthermore, online detection frameworks have addressed specific calibration issues, such as extrinsic spatial drift induced by vibration and thermal stress~\cite{Peng2024Miscalibration,Xia2025Misalignment,Tahiraj2025CalorNoCal}.

Despite these advances, FDD development remains constrained by the absence of a unified, cross-modal fault taxonomy. Most diagnostic models are trained on datasets heavily biased toward adverse weather~\cite{BoreasDataset,TWICEdataset,SIDdataset,DAWNdataset,WADSdataset}, while true hardware and logical faults remain underrepresented. Although recent efforts have injected synthetic sensor anomalies (e.g., LiDAR dropouts) into existing datasets~\cite{MultiModalExpert,BEVFusion}, these interventions are fragmented. The lack of a comprehensive framework that traces faults from their physical origins to their perceptual consequences limits the systematic evaluation and generalization of both sensor fusion and diagnostic models.

\subsection{Existing Surveys and Positioning of This Work}
\label{sec:positioning}

While numerous studies have investigated isolated sensor degradations or evaluated downstream algorithmic robustness, efforts to unify these phenomena into a cross-modal framework remain scarce. 

\begin{table}[!t]
\caption{Comparison of Representative Sensor Fault Surveys, Benchmarks, and Frameworks.}
\label{tab:survey_comparison}
\centering
\scriptsize
\setlength{\tabcolsep}{3pt}
\resizebox{\columnwidth}{!}{%
\begin{tabular}{lccccccc}
\toprule
\textbf{Work} & \textbf{Cam.} & \textbf{LiDAR} & \textbf{Radar} & \textbf{Subsys.} & \textbf{Soft./Cal.} & \textbf{Synerg.} & \textbf{Toolkit} \\
\midrule
Secci~\cite{RGB_CameraFailures} & \checkmark & -- & -- & \checkmark & \checkmark & -- & \checkmark \\
Chang et al.~\cite{Chang2023}        & -- & \checkmark & -- & \checkmark & Partial & -- & -- \\
Dong et al.~\cite{Dong2023}          & \checkmark & \checkmark & -- & -- & Partial & Partial & \checkmark \\
FADE~\cite{FADE2025}                 & \checkmark & \checkmark & -- & \checkmark & Partial & Partial & -- \\
Jin et al.~\cite{Baseline}           & \checkmark & \checkmark & -- & -- & Partial & -- & -- \\
Robo3D~\cite{kong2023robo3d}         & -- & \checkmark & -- & -- & -- & -- & \checkmark \\
Matos et al.~\cite{Matos2025SimulatingSensorFailures} & -- & \checkmark & -- & -- & Partial & -- & \checkmark \\
Matos et al. (survey)~\cite{matos2024survey} & \checkmark & \checkmark & \checkmark & Partial & Partial & -- & -- \\
Viktor~\cite{viktor2026modality}     & \checkmark & \checkmark & \checkmark & Partial & Partial & Partial & -- \\
\midrule
\textbf{Ours}                        & \checkmark & \checkmark & \checkmark & \checkmark & \checkmark & \checkmark & \checkmark \\
\bottomrule
\end{tabular}%
}
\end{table}

Table~\ref{tab:survey_comparison} contrasts these works along the modalities they cover and the aspects of the taxonomy they address. Component-level analyses exist for cameras~\cite{RGB_CameraFailures} and LiDARs~\cite{Chang2023}, and FADE~\cite{FADE2025} combines both in active (internal damage) and passive (external) fault models within a fuzzing-based testing framework. Dong et al.~\cite{Dong2023} and Robo3D~\cite{kong2023robo3d} provide corruption benchmarks without mapping faults to subsystems, while Matos et al.~\cite{Matos2025SimulatingSensorFailures} inject silent failures and noise into LiDAR, GNSS, and IMU data in CARLA and Autoware. Jin et al.~\cite{Baseline} use the radar only as a fault-free reference to diagnose corruptions and alignment errors injected into the camera and LiDAR streams. Compound faults are considered only marginally: FADE injects co-faults caused by the same environmental condition~\cite{FADE2025}, and Dong et al. test concurrent LiDAR and camera noise~\cite{Dong2023}. General-purpose fault injection tools for autonomous vehicles target the driving stack rather than the sensor faults themselves: AVFI~\cite{jha2018avfi} injects faults into sensor inputs (camera or LiDAR fault models), neural networks, and hardware or software components within CARLA, DriveFI~\cite{jha2019ml} mines the faults that maximally impact safety on industry-grade driving stacks, and CarFASE~\cite{maleki2023carfase} is an open-source CARLA-based fault and attack simulation engine demonstrated on OpenPilot with brightness and salt-and-pepper faults.

Two recent surveys are closest in scope. Matos et al.~\cite{matos2024survey} review publications on sensor failures in autonomous vehicles, covering also ultrasonic sensors, GNSS, and IMUs, and discuss mitigation strategies such as sensor fusion, redundancy, and calibration. Viktor~\cite{viktor2026modality} conducts a PRISMA-based systematic review of sensor architectures, introducing a modality-to-safety taxonomy that links sensing principles to degradation mechanisms and vehicle-level consequences while distinguishing sensor degradation, sensor failure, and perception failure, and proposes a reliability-aware architecture.

Although both surveys offer valuable system-level perspectives, three gaps remain:
\begin{enumerate}
    \item \textit{Modality- or outcome-driven structure:} existing reviews organize anomalies predominantly by individual sensor or by safety outcome, and the closest origin-based classification, the active and passive faults of FADE~\cite{FADE2025}, is limited to camera and LiDAR; a classification that groups camera, LiDAR, and radar faults simultaneously by their root cause is missing.
    \item \textit{Partial coverage of embedded software and compound faults:} pre-inference firmware, ISP, and DSP anomalies are treated for single modalities, such as the ISP faults of cameras~\cite{RGB_CameraFailures}, and the interaction of concurrent heterogeneous faults is, to the best of our knowledge, not systematically cataloged.
    \item \textit{Limited operational support:} injection tools exist, but they cover a single modality~\cite{RGB_CameraFailures,kong2023robo3d}, at most camera and LiDAR~\cite{Dong2023,FADE2025}, LiDAR, GNSS, and IMU~\cite{Matos2025SimulatingSensorFailures}, or target the driving stack rather than a sensor-level fault taxonomy~\cite{jha2018avfi,jha2019ml,maleki2023carfase}; none ties its fault list to a cross-modal taxonomy.
\end{enumerate}

The framework proposed in this paper addresses these gaps. It establishes a cross-modal, origin-driven taxonomy across camera, LiDAR, and radar, maps each fault to its subsystem or processing layer, identifies synergistic fault couplings, and implements the classification in an open-source fault injection toolkit (94 routines) with graded severities and reproducible frame-level annotations.

\section{Proposed Sensor Fault Taxonomy}
\label{sec:taxonomy}

This section introduces a unified, cross-modal taxonomy of faults affecting exteroceptive sensors in autonomous vehicles. Unlike classifications organized by individual sensor or by vehicle-level safety outcome, the proposed framework groups camera, LiDAR, and radar anomalies by their root cause and systematically relates them to their observable symptoms, affected subsystems, and direct impact on environmental perception. The scope is limited to camera, LiDAR, and radar, the primary exteroceptive suite for safety-critical 3D scene understanding in modern AVs, which are also available in the Boreas dataset used in Section~\ref{sec:simulations}; other exteroceptive sensors, such as ultrasonic sensors, are not covered.

Building on the functional safety terminology introduced in Section~\ref{sec:introduction}, the taxonomy rests on three guiding principles: 1)~\textit{cross-modality consistency}, defining fault categories independently of the specific sensor technology; 2)~\textit{architectural and perceptual relevance}, linking each anomaly to the hardware or firmware subsystem where it originates and to the resulting distortion of the sensor output; and 3)~\textit{safety traceability}, explicitly separating mechanisms to support distinct safety analyses. Specifically, hardware, calibration, and software anomalies align with the strict ISO~26262 definition of a fault, whereas environmental interference corresponds to performance limitations and triggering conditions under the SOTIF (ISO~21448) standard~\cite{ISO26262_1,ISO21448_2022}. We use the term \emph{fault} broadly to encompass both, as both corrupt the perception output, and synergistic faults frequently involve their interaction.

Following these principles, the framework organizes exteroceptive sensor faults into five principal categories:

\begin{itemize}
    \item \textbf{Hardware Faults:} Physical degradations, component aging, or electrical malfunctions occurring within the sensor’s structural, optical, or electronic assemblies.
    \item \textbf{Environmental Faults:} External phenomena that interfere with signal propagation or acquisition, or stress the sensor hardware from outside.
    \item \textbf{Calibration Faults:} Inaccuracies or drifts in the intrinsic, extrinsic, or radiometric parameters that govern sensor alignment and image formation.
    \item \textbf{Software Faults:} Logical errors, buffer overflows, or driver instability within the sensor’s embedded firmware, ISP, or DSP, strictly prior to downstream AI inference.
    \item \textbf{Synergistic Faults:} Emergent, non-linear fault conditions arising from the simultaneous interaction of multiple heterogeneous fault sources.
\end{itemize}

Hardware, environmental, and software faults follow the signal chain, from physical generation and acquisition, through propagation in the external medium, to embedded digital processing. Calibration faults concern the geometric and radiometric parameters of the sensors and their spatial relation to the vehicle frame, while synergistic faults combine causes from the preceding categories. A fault is assigned to the category of its root cause; the same physical phenomenon may appear in two categories if it acts through different mechanisms (e.g., thermal stress causing structural hardware deformation versus long-term calibration drift).

Each entry in the taxonomy is characterized by four dimensions: its root-cause category, its physical or logical mechanism and observable symptom, the affected sensing modality, and the target subsystem or processing layer. Temporal behavior (permanent, intermittent, or transient, following the ISO~26262 classification) and impact on perception are discussed qualitatively within each category, and are operationalized in the toolkit of Section~\ref{sec:simulations} through persistent and transient fault injection patterns.

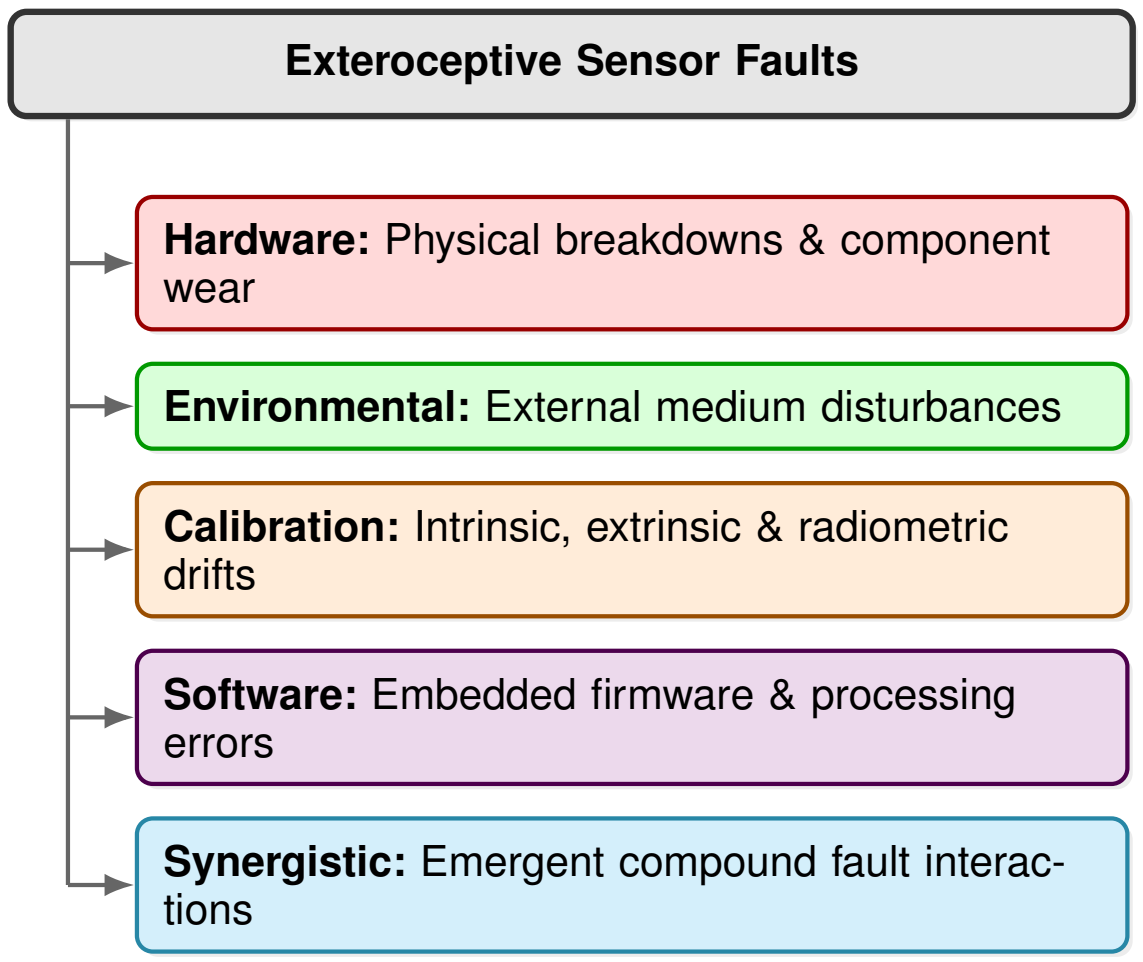
\begin{figure}[h!]
\setlength{\tabcolsep}{3.5pt}
\centering
\begin{tikzpicture}[
  font=\sffamily\footnotesize,
  % Stile per i blocchi delle categorie
  category/.style={
    draw, 
    rectangle, 
    rounded corners=3pt, 
    thick, 
    inner sep=5pt, 
    text width=6.2cm, 
    align=left,
    drop shadow={opacity=0.15, shadow xshift=1pt, shadow yshift=-1pt}
  },
  % Stile per le linee e le frecce
  line/.style={draw, -{Latex[length=2mm,width=1.5mm]}, thick, draw=black!60}
]

% Nodo Radice (Root)
\node[draw=black!80, fill=gray!20, rectangle, rounded corners=3pt, very thick, inner sep=6pt, text width=7cm, align=center, drop shadow={opacity=0.15, shadow xshift=1pt, shadow yshift=-1pt}] (root) {\textbf{Exteroceptive Sensor Faults}};

% Nodi Categorie (impilati verticalmente, leggermente spostati a destra)
\node[category, fill=red!15, draw=red!60!black, below=0.5cm of root, xshift=0.4cm] (hard) {\textbf{Hardware:} Physical breakdowns \& component wear};
\node[category, fill=green!15, draw=green!60!black, below=0.2cm of hard] (env) {\textbf{Environmental:} External medium disturbances};
\node[category, fill=orange!15, draw=orange!60!black, below=0.2cm of env] (cal) {\textbf{Calibration:} Intrinsic, extrinsic \& radiometric drifts};
\node[category, fill=violet!15, draw=violet!60!black, below=0.2cm of cal] (soft) {\textbf{Software:} Embedded firmware \& processing errors};
\node[category, fill=cyan!15, draw=cyan!60!black, below=0.2cm of soft] (syn) {\textbf{Synergistic:} Emergent compound fault interactions};

% Disegno la linea dorsale (Spine) e le diramazioni
\draw[thick, draw=black!60] (root.south -| root.west) ++(0.4cm,0) coordinate (spineTop) -- (spineTop |- syn.west) coordinate (spineBot);

\draw[line] (spineTop |- hard.west) -- (hard.west);
\draw[line] (spineTop |- env.west) -- (env.west);
\draw[line] (spineTop |- cal.west) -- (cal.west);
\draw[line] (spineTop |- soft.west) -- (soft.west);
\draw[line] (spineTop |- syn.west) -- (syn.west);

\end{tikzpicture}
\caption{Classification of exteroceptive sensor faults.}
\label{fig:sensor_faults}
\end{figure}

As illustrated in Fig.~\ref{fig:sensor_faults} and synthesized across modalities in Table~\ref{tab:comprehensive_faults_taxonomy}, this classification establishes a common vocabulary for fault modeling, facilitating systematic fault injection and the evaluation of robust perception algorithms.

\begin{table*}[p]
\centering
\scriptsize
\caption{Cross-Modal Taxonomy of Exteroceptive Sensor Faults: Causes, Symptoms, and Subsystem Mapping.}
\label{tab:comprehensive_faults_taxonomy}
\renewcommand{\arraystretch}{1.06}
\setlength{\tabcolsep}{3.5pt}
\begin{tabular}{p{0.21\textwidth} p{0.535\textwidth} p{0.095\textwidth} p{0.105\textwidth}}
\hline\hline
\textbf{Fault Name} & \textbf{Physical / Logical Description} & \textbf{Affected Sensor} & \textbf{Subsystem / Layer} \\
\hline

\rowcolor{red!15}
\multicolumn{4}{l}{\textbf{CATEGORY I: HARDWARE FAULTS}} \\
\hline
Brightness~\cite{RGB_CameraFailures} & Failures in lens mechanics (shutter, diaphragm) compromising light transmission. & Camera & Lens \\
Broken Lens~\cite{RGB_CameraFailures,FADE2025} & Impact damage causing linear fractures, scattering, or irregular image artifacts. & Camera & Lens \\
Broken VR~\cite{RGB_CameraFailures} & Failure of the vibration reduction system leading to out-of-focus or blurred images. & Camera & Lens \\
Heat Failure~\cite{RGB_CameraFailures} & Excessive temperature evaporating lubricants, impairing autofocus or zoom mechanics. & Camera & Lens / Body \\
Electrical Overload~\cite{RGB_CameraFailures} & Overheating and electronic damage causing corrupted images or complete signal loss. & Camera & Body \\
No Bayer Filter~\cite{RGB_CameraFailures} & Failure of the Bayer filter operation causing inaccurate color reconstruction. & Camera & Bayer Filter \\
Banding~\cite{RGB_CameraFailures} & Parallel horizontal or vertical lines visible across the image, originating in the image sensor. & Camera & Image Sensor \\
Bright Lines~\cite{RGB_CameraFailures} & Extreme external optical energy (e.g., LiDAR lasers) permanently damaging pixels. & Camera & Image Sensor \\
Dead Pixel~\cite{RGB_CameraFailures,OnlineImageFaults} & Individual pixels failing to respond, producing static dark or bright dots. & Camera & Image Sensor \\
No Action~\cite{RGB_CameraFailures} & ISP hardware failure preventing raw-to-RGB conversion pipeline execution. & Camera & ISP \\
Laser Device Fault~\cite{Chang2023,FADE2025} & Laser diode degradation lowering optical power and reducing detection range. & LiDAR & Emitter \\
Photodetector Fault~\cite{Chang2023} & Reduced sensitivity or increased noise in the photodetector, introducing spurious points. & LiDAR & Detector \\
Microelectronics Fault~\cite{Chang2023} & Aging in timing/control circuits degrading synchronization, dropping frames, or halting output. & LiDAR & Electronics \\
Electrical Conn. Fault~\cite{Chang2023} & Vibration-induced loosening of connectors/splices interrupting power or data lines. & LiDAR & Interconnects \\
Scanning Mech. Fault~\cite{Chang2023} & Wear in rotating (and, to a lesser extent, MEMS) scanners reducing scan rate, field of view, and resolution. & LiDAR & Scanner \\
Optical System Fault~\cite{Chang2023} & Vibration-induced lens or mirror misalignment reducing FOV and missing detections. & LiDAR & Optics \\
Thermal Mgmt. Fault~\cite{Chang2023} & Degraded thermal contact/paste raising temperature, lowering power and adding noise. & LiDAR & Cooling \\
Enclosure Fault~\cite{Chang2023} & Seal degradation allowing dust/moisture ingress, leading to internal optical scattering. & LiDAR & Housing \\
LNA Degradation~\cite{RadarGracefulDegradation} & Gain degradation of the receiver low-noise amplifier reducing detection performance. & Radar & Front-End \\
Antenna Array Failure~\cite{RadarGracefulDegradation} & Failure of array elements altering the radiation pattern and degrading gain and angular accuracy. & Radar & Antenna \\
Thermal Deformation~\cite{Albrecht2022} &
Thermal stress causing deformation of radar packaging components. &
Radar & Packaging \\
\hline

\rowcolor{green!15}
\multicolumn{4}{l}{\textbf{CATEGORY II: ENVIRONMENTAL FAULTS}} \\
\hline
Rain~\cite{RGB_CameraFailures,FADE2025,Dong2023,Rob_LidarBased,kong2023robo3d,Radar_Sediments,Dust_Radar} & Water accumulation causing optical scattering, RF attenuation, and radome losses. & All & Lens / Radome \\
Snow / Ice~\cite{RGB_CameraFailures,FADE2025,Dong2023,Rob_LidarBased,kong2023robo3d,LiDAR_snow,Radar_Sediments} & Solid accumulation causing absorption, backscattering, and antenna detuning. & All & Lens / Radome \\
Fog~\cite{sensorsFog,FADE2025,Dong2023,kong2023robo3d,LiDAR_fog} & Volumetric scattering attenuating optical/laser signals, reducing contrast and range. & Cam / LiDAR & Optical Path \\
External Scatter~\cite{RGB_CameraFailures,FADE2025,kong2023robo3d,LiDAR_Dust,Dust_Radar} & Airborne dust, smoke, or debris causing contrast loss, spurious points, and clutter. & All & Path / Radome \\
Window Contamination~\cite{FADE2025,Chang2023,kong2023robo3d,RGB_CameraFailures} & Mud, dirt, or insects on protective surfaces causing blind spots or attenuation. & All & Lens / Radome \\
Strong Sunlight / Flare~\cite{RGB_CameraFailures,FADE2025,Dong2023} & Direct light or reflections causing sensor saturation, lowering SNR and missing targets. & Cam / LiDAR & Lens / Detector \\
Condensation~\cite{FADE2025,RGB_CameraFailures,Dong2023,Radar_Sediments,Dust_Radar} & Humidity-induced moisture film causing optical blur and RF signal detuning. & Cam / Radar & Lens / Radome \\
Interference / Crosstalk~\cite{FADE2025,Baseline,kong2023robo3d,LiDAR_Crosstalk,Radar_Artifacts,Radar_anomaly} & Unsynchronized active emitters (other LiDARs/radars) introducing structured noise. & LiDAR / Radar & Receiver \\
Spots~\cite{RGB_CameraFailures} & Microscopic dust settling on the image sensor and creating static dark spots in the image. & Camera & Image Sensor \\
Brackish Corrosion~\cite{RGB_CameraFailures,OutdoorCorrosion} & Salt and humidity corroding mechanical seals, facilitating internal ingress. & All & Housing \\
Water Ingress~\cite{RGB_CameraFailures} & Liquid penetration into the sensor housing causing short-circuits and total failure. & All & Housing \\
Wet Ground~\cite{kong2023robo3d,LiDAR_Wet} & Water layers on asphalt absorbing laser energy, causing missing ground returns. & LiDAR & Optical Path \\
Incomplete Echo~\cite{kong2023robo3d} & Dark or absorptive materials reflecting insufficient laser energy to the detector. & LiDAR & Detector \\
Mechanical Instability~\cite{RGB_CameraFailures,kong2023robo3d,Chang2023} & Vehicle vibrations and mounting loosening causing dynamic pitch/roll variations. & Cam / LiDAR & Mounting \\
Multipath~\cite{Radar_Artifacts,Radar_anomaly} & Indirect RF reflections from infrastructure generating delayed echoes (ghost targets). & Radar & Path \\
Clutter~\cite{Radar_Artifacts} & Unwanted backscattering from the road surface or rain obscuring legitimate targets. & Radar & Path \\
Foliage Interference~\cite{ITU833} & Signal absorption and scattering caused by roadside vegetation. & Radar & Path \\
\hline

\rowcolor{orange!15}
\multicolumn{4}{l}{\textbf{CATEGORY III: CALIBRATION FAULTS}} \\
\hline
Blurred Image~\cite{RGB_CameraFailures,MotionBlurRecovery} & Loss of optical sharpness due to intrinsic focal length drift or autofocus error. & Camera & Lens \\
White Balance Error~\cite{FADE2025} & Incorrect color gain calibration leading to severe chromatic bias (cool/warm tints). & Camera & ISP \\
Lens Distortion Error~\cite{Peng2024Miscalibration,RGB_CameraFailures} & Inaccurate geometric correction parameters for fisheye or wide-angle models. & Camera & ISP \\
Boresight Misalignment~\cite{RadarMisalignment} & Azimuth/elevation angular offset between the antenna RF axis and physical mounting. & Radar & Antenna \\
Thermal Expansion Drift~\cite{Kettelgerdes2026,HarterRadar} &
Temperature induced drift of sensor calibration parameters, affecting measurement accuracy. &
LiDAR / Radar & Calibration \\
Extrinsic Misalignment~\cite{Peng2024Miscalibration,Xia2025Misalignment,Dong2023,Baseline,RGB_CameraFailures,Tahiraj2025CalorNoCal,Chang2023} & Mechanical vibrations shifting the relative 6-DoF transformation between sensors. & All & Calibration \\
\hline

\rowcolor{violet!15}
\multicolumn{4}{l}{\textbf{CATEGORY IV: SOFTWARE FAULTS}} \\
\hline
Aberration Fail~\cite{RGB_CameraFailures,ChromaticAberration} & ISP firmware failure to compensate for wavelength-dependent lens dispersion. & Camera & ISP \\
Incomplete Demosaicing~\cite{RGB_CameraFailures} & Failure in reconstructing RGB values from raw Bayer data, causing pixelated images. & Camera & ISP \\
Incomplete Noise Reduct.~\cite{RGB_CameraFailures} & Partial execution of spatial/temporal denoising algorithms increasing image noise. & Camera & ISP \\
Interference Mitig. Fault~\cite{Radar_software} & Failure of on-board DSP algorithms to suppress external RF interference. & Radar & DSP \\
Inference Timing Fault~\cite{Radar_software2} & Delays in on-board processing producing outdated target lists. & Radar & DSP \\
Temporal Misalignment~\cite{Dong2023,Chang2023,Baseline} & Latency or desynchronization in timestamping routines degrading temporal fusion. & All & Firmware \\
Silent Failure~\cite{Matos2025SimulatingSensorFailures} & Embedded driver or communication interface crash halting data transmission. & All & Comm. I/F \\
Data Handling Fault~\cite{Dong2023,Baseline} & Software-level packet loss or frame misordering within the acquisition pipeline. & All & Firmware \\
Internal Filtering Fault\cite{Giuffrida2023,Rao2025} & Erroneous firmware thresholding or filtering discarding valid detections. & LiDAR / Radar & DSP / Filter \\
\hline

\rowcolor{cyan!15}
\multicolumn{4}{l}{\textbf{CATEGORY V: SYNERGISTIC FAULTS}} \\
\hline
Broken Lens + Condensation & A fracture facilitating moisture ingress; its interaction with condensation causes complex scattering and refraction. & Camera & Lens \\
Scanning Fault + Fog & Scanner instability combined with aerosol scattering creating irregular, dynamic blind spots. & LiDAR & Scanner / Path \\
Thermal Deform. + Multipath &
Thermally induced packaging deformation altering beam direction, compounding ghost targets. &
Radar & Packaging / Path \\
Extrinsic Misalign. + Fog & Spatial coordinate drift compounded by fog attenuation, breaking spatial fusion. & Cam / LiDAR & Calib. / Path \\
Temporal Misalign. + Multipath & Frame desynchronization interacting with delayed radar echoes (radar) or frame blending (camera) to duplicate targets. & Cam / Radar & Sync. / Path \\
Calibration Drift + Interference & Thermal range or mounting drift combined with LiDAR crosstalk or radar RF interference. & LiDAR / Radar & Calib. / Rx \\
\hline\hline
\end{tabular}
\end{table*}

\subsection{Hardware Faults}
\label{sec:hardware_faults}

Hardware faults represent physical degradations or malfunctions that directly affect the sensing unit or its associated electronic and optical components. Altering the raw signal at its origin, these faults are typically permanent or progressively worsening, and induce measurement bias, signal instability, or complete data loss, severely compromising perception accuracy. LiDAR failure modes, for instance, are classified as sudden failures or slow degradation~\cite{Chang2023}. A detailed mapping of these faults is provided in Category I of Table~\ref{tab:comprehensive_faults_taxonomy}.

\textbf{Cameras} suffer from permanent faults affecting image acquisition at its source. Typical examples involve damaged lenses \cite{FADE2025}, defective vibration reduction mechanisms, electrical overloads \cite{RGB_CameraFailures}, or sensor-level degradations such as dead pixels \cite{OnlineImageFaults}, banding, and color reconstruction errors within the Bayer filter \cite{RGB_CameraFailures}.

\textbf{LiDAR} systems, relying on the coordinated performance of optical, electronic, and mechanical subsystems, experience hardware degradations primarily due to component aging and mechanical wear \cite{Chang2023,FADE2025}. Laser diode degradation and photodetector sensitivity loss reduce the detection range and add noise, wear in the scanning mechanism (mainly in rotating scanners, less in MEMS-based ones) reduces scan rate, field of view, and resolution, and failures of the timing electronics cause dropped frames \cite{Chang2023}.

\textbf{Radar} sensors without moving parts are mechanically simpler, and radar hardware faults are comparatively less documented. They typically involve gradual electrical degradation (e.g., LNA gain reduction) or localized antenna element failures in array radars \cite{RadarGracefulDegradation}; mechanically rotating radars are an exception. Additionally, inadequate thermal management can compromise radar performance and reliability \cite{Thermal_man_Radar}.

\subsection{Environmental Faults}
\label{sec:env_faults}

Environmental faults originate from external conditions that interfere with signal propagation or acquisition, or that stress the sensor from outside. Typical sources include adverse weather (rain, fog, snow), contamination and corrosion, and optical or electromagnetic disturbances (glare, crosstalk, multipath reflections), which attenuate signals or generate false detections. These context-dependent faults are cataloged in Category II of Table~\ref{tab:comprehensive_faults_taxonomy}.

\textbf{Cameras} operate within the visible spectrum and are extremely sensitive to variations in illumination, contrast, and visibility. Environmental faults manifest as glare, low sun angles, lens condensation, occlusions from raindrops, and contrast loss in fog \cite{RGB_CameraFailures, FADE2025, Dong2023}. While often transient, they severely degrade feature extraction and dynamic range.

\textbf{LiDAR} systems, while independent of ambient illumination, remain vulnerable to atmospheric scattering and attenuation caused by fog, rain, or snow, leading to reduced range and point density \cite{Dong2023,LiDAR_fog,LiDAR_snow}. Contaminants on the optical window (dirt, mud, ice) obstruct the optical path \cite{FADE2025, Chang2023}, while highly absorptive surfaces and crosstalk from other LiDARs introduce structured noise and missing echoes \cite{kong2023robo3d,LiDAR_Crosstalk}.

\textbf{Radar} is intrinsically resilient to poor visibility and optical scattering \cite{sensorsFog}. However, RF signals interact with the environment, leading to performance degradation under specific conditions. Contaminants accumulated on the radome attenuate transmissions \cite{Radar_Sediments, Dust_Radar}, while highly reflective surroundings generate multipath reflections, clutter, and ghost targets \cite{Radar_Artifacts, Radar_anomaly}, causing false obstacle detections.

\subsection{Calibration Faults}
\label{sec:calibration_faults}

Calibration faults arise from inaccuracies or temporal drifts in the intrinsic, extrinsic, or radiometric parameters governing sensor alignment and image formation \cite{Peng2024Miscalibration, Xia2025Misalignment}. Driven by environmental stress, mechanical vibration, or aging, these faults distort spatial transformations, leading to inconsistent multi-sensor fusion outputs, as outlined in Category III of Table~\ref{tab:comprehensive_faults_taxonomy}. Timing offsets between sensors are cataloged as software faults, since they stem from latency or timestamping in the firmware. Calibration faults are typically permanent: gradual when produced by aging or vibration~\cite{Chang2023}, and abrupt when caused by a shock or a repair. Gradual drifts are particularly insidious because the system keeps operating while the consistency between sensors deteriorates incrementally~\cite{viktor2026modality}.

\textbf{Cameras} experience intrinsic errors involving deviations in focal length, principal point, or lens distortion, resulting in geometric warping or image blur \cite{RGB_CameraFailures, MotionBlurRecovery}. Radiometric miscalibrations (incorrect white balance) alter color constancy. Extrinsic faults manifest as physical misalignments relative to the vehicle frame, degrading cross-sensor registration \cite{Peng2024Miscalibration, Xia2025Misalignment}.

\textbf{LiDAR} systems suffer from range biases that distort point distributions. Extrinsic errors offset the generated 3D point cloud with respect to the vehicle frame, significantly compromising object detection boundaries and pose estimation algorithms \cite{Dong2023, Baseline, Chang2023}.

\textbf{Radar} calibration faults primarily manifest as spatial biases affecting beam orientation. Boresight misalignments in azimuth, elevation, or roll generate consistent directional errors \cite{RadarMisalignment}. Temperature variations can also induce drift in radar calibration parameters, potentially affecting angular measurement accuracy~\cite{HarterRadar}.

\subsection{Software Faults}
\label{sec:soft_faults}

Software faults are defined strictly as anomalies within the sensor’s embedded firmware, internal processing layer, or manufacturer-provided SDK, explicitly excluding downstream AI inference errors. Stemming from logical inconsistencies, buffer issues, or timing routines, these faults produce desynchronization, packet loss, or corrupted formatting without necessarily causing visible physical artifacts (Category IV of Table~\ref{tab:comprehensive_faults_taxonomy}). Their temporal behavior ranges from transient or intermittent, as for packet loss, frame drops, or timing jitter, to permanent, as for a driver crash that halts data transmission until the sensor or process is restarted.

\textbf{Camera} software faults originate within the Image Signal Processor (ISP) \cite{RGB_CameraFailures}. Failures in digital routines such as demosaicing, chromatic aberration correction, or noise reduction yield pixelated, color-shifted, or degraded outputs before the data reaches the perception network. Additionally, frame drops and timestamping errors disrupt temporal fusion \cite{Dong2023,Baseline}.

\textbf{LiDAR} firmware anomalies manifest in data-handling pipelines. Packet loss and frame misordering result in missing point cloud sectors or temporal shifts \cite{Dong2023,Baseline}, and erroneous range gating or thresholding in the internal filtering stack, which the toolkit models, discards valid returns. Communication-layer interruptions can lead to silent failures, halting data transmission entirely \cite{Matos2025SimulatingSensorFailures}.

\textbf{Radar} software faults affect the embedded Digital Signal Processor (DSP). Interference from other radars corrupts range-Doppler maps and generates false targets, and the on-board mitigation stage \cite{Radar_software} must suppress it; its failure leaves the interference unmitigated. Delays in on-board processing, which become relevant on embedded platforms with adaptive processing \cite{Radar_software2}, produce outdated target lists and break the temporal coherence critical for dynamic obstacle tracking.

\subsection{Synergistic Faults}
\label{sec:syner_faults}

Synergistic faults occur when multiple heterogeneous fault sources (e.g., hardware degradation combined with environmental interference) interact simultaneously. These combinations can generate non-linear, emergent behaviors that invalidate basic sensor-fusion assumptions. Because their symptoms may mimic natural scene variability or complex clutter, they can bypass standard Fault Detection and Diagnosis (FDD) thresholds (Category V of Table~\ref{tab:comprehensive_faults_taxonomy}). The combinations below are derived by combining the single faults of the previous categories. Their dynamics combine those of their components: a permanent condition, such as a damaged lens or a deformed package, modulated by transient effects, such as fog density or multipath geometry, which the toolkit of Section~\ref{sec:toolkit} reproduces by combining a burst-level pattern with per-frame variations.

\textbf{Cameras} can experience optical-environmental couplings. For example, a minor mechanical fracture in the lens facilitates moisture ingress. The interaction between the crack geometry and condensation dynamics can produce complex light scattering and refraction, yielding severe and dynamic image distortion.

\textbf{LiDAR} systems are susceptible to mechanical-environmental synergies. An imbalance in the scanning mechanism (hardware), coupled with atmospheric scattering from fog (environmental), can generate an irregular point-cloud density with fluctuating blind spots that are hard to separate from natural scene variability.

\textbf{Radar} measurements can degrade non-linearly under structural-electromagnetic synergies. Thermal gradients generating minor packaging deformation (hardware) can alter the antenna beam direction. In reflective environments, this angular shift interacts with multipath propagation (environmental), producing temperature-dependent ghost targets and unstable target positions.

\section{Simulated Examples of Sensor Faults}
\label{sec:simulations}

To visually demonstrate the perceptual impact of the taxonomy categories defined in Section~\ref{sec:taxonomy}, a set of representative sensor faults was injected into the Boreas dataset~\cite{BoreasDataset}. The Boreas platform provides high-quality, multi-modal data in diverse conditions, serving as a suitable baseline for fault simulation and validation. The reference sensor suite employed in this dataset includes:

\begin{itemize}
    \item \textbf{LiDAR:} Velodyne Alpha Prime (VLS-128) with 128 beams, an angular resolution of $0.1^{\circ}$ vertically and $0.2^{\circ}$ horizontally, a $360^{\circ} \times 40^{\circ}$ field of view, a maximum range of 300~m, and an operating frequency of 10~Hz.
    \item \textbf{Radar:} Navtech CIR304-H with a range resolution of 0.0596~m and a maximum range of about 200~m for the sequences recorded before a firmware upgrade of the sensor (0.0438~m and 250~m afterwards), an angular resolution of $0.9^{\circ}$, and an operating frequency of 4~Hz. The radar frame used in this work belongs to the earlier configuration, with 400 azimuths and 3360 range bins.
    \item \textbf{Camera:} FLIR Blackfly S (BFS-U3-51S5C) RGB camera with a resolution of 2448~$\times$~2048~pixels (5~MP), a field of view of $81^{\circ} \times 71^{\circ}$, and a frame rate of 10~Hz.
\end{itemize}

\subsection{Open-Source Fault Injection Toolkit}
\label{sec:toolkit}

To support reproducibility and systematic benchmarking, the fault injection procedures used to generate the examples in this section are released as an open-source toolkit. The toolkit realizes the taxonomy of Section~\ref{sec:taxonomy} in software: each fault entry of Table~\ref{tab:comprehensive_faults_taxonomy} that applies to a given sensor is implemented as a dedicated injection routine, and the routines of each modality are organized in five modules, one per fault category (hardware, environmental, calibration, software, and synergistic). Camera, LiDAR, and radar streams are handled independently and operate directly on the native formats of the Boreas dataset (images, binary point clouds, and polar radar images); supporting another dataset only requires a loader for its data format.

\subsubsection{Architecture and Taxonomy Mapping}
Each modality is driven by an orchestrator that dispatches every frame to the category modules through a registry of canonical keys of the form \texttt{category:fault}. The registry maps each routine to its taxonomy entry and to the affected subsystem (e.g., lens, body, ISP, or image sensor for cameras; emitter, detector, scanner, or optics for LiDAR; front-end, antenna, housing, radome, or DSP for radar), so that every corrupted frame is annotated with the entry and the subsystem it originates from. Table~\ref{tab:toolkit_coverage} summarizes the number of routines per category and modality. Some taxonomy entries are realized by several modes (e.g., total or partial electrical overload), and some routines cover more than one entry when the observable symptom coincides (e.g., rain, fog, and external scatter in the LiDAR atmospheric scattering model). Water ingress, brackish corrosion, and the antenna array failure are modeled as functional proxies that reproduce their effect on the sensor output. 

\begin{table}[!t]
\caption{Number of Injection Routines per Taxonomy Category and Sensor Modality}
\label{tab:toolkit_coverage}
\centering
\begin{tabular}{lccc}
\toprule
Category & Camera & LiDAR & Radar \\
\midrule
Hardware      & 15 & 8  & 3  \\
Environmental & 13 & 10 & 11 \\
Calibration   & 4  & 2  & 3  \\
Software      & 6  & 4  & 6  \\
Synergistic   & 3  & 3  & 3  \\
\midrule
Total         & 41 & 27 & 26 \\
\bottomrule
\end{tabular}
\end{table}

\subsubsection{Severity Model and Fault Scheduling}
Every fault with a graded effect exposes four severity levels (\textit{mild}, \textit{moderate}, \textit{severe}, \textit{extreme}), each associated with a preset of the fault parameters; binary faults, such as a silent driver failure, have no severity scale and are labeled as not applicable. Where a physical relation is available, it is used to structure the presets. For example, the LiDAR atmospheric scattering and snow models attenuate the return intensities by $\exp(-2\sigma r)$ (two-way Beer--Lambert law) and discard the returns beyond the meteorological visibility $V = 3.912/\sigma$, given by the Koschmieder relation with a 2\% contrast threshold~\cite{lee2016visibility,horvath1971applicability}; the extinction coefficient $\sigma$ increases with severity, and the relation is used here as a simple severity parametrization rather than as a model of the LiDAR detection range. Similarly, the radar rain routine models two-way path attenuation with the ITU-R P.838-3 power law~\cite{ITU838}, $\gamma_R = k\,R^{\alpha}$, where $\gamma_R$ is the specific attenuation in dB/km, $R$ the rain rate in mm/h, and $k$ and $\alpha$ depend on frequency and polarization, whereas antenna array failures apply a coherent gain loss of $-20\log_{10}(1-f)$~dB, a function of the fraction $f$ of failed elements. The remaining presets are heuristic: the strength of each parameter increases with the severity level.

Faults are scheduled as bursts, i.e., intervals of contiguous frames during which a fault is active, which models temporally extended (permanent or intermittent) faults. A burst can be specified explicitly in a JSON plan or generated randomly from a number of bursts, a burst-length range, a target fraction of faulty frames, and a severity policy (fixed, cyclic, or random with user-defined weights); the random generation stops when either the target fraction or the number of bursts is reached. For faults that physically degrade gradually (e.g., LiDAR laser aging, radar LNA aging, antenna failure, or thermal drift), bursts can also be configured, in random mode, to ramp progressively from \textit{mild} to \textit{extreme}, each severity segment being recorded as a burst of its own. Each frame receives at most one fault, and a configurable number of leading and trailing frames is kept clean to provide fault-free context.

\subsubsection{Ground-Truth Annotation and Reproducibility}
Faults that model a permanent physical condition, such as a crack in the lens, a deposit on the protective window or radome, a thermal packaging deformation, or a mounting misalignment, are fixed-pattern: they keep the same random pattern throughout a burst. Faults that are inherently transient are frame-varying, drawing a new pattern for every frame, and synergistic faults, as well as some environmental ones, combine a fixed-pattern component with a frame-varying one. All random choices derive from a single master seed.

Faults that depend on other frames of the sequence are handled at the sequence level. For cameras, a silent failure repeats the last healthy frame preceding the burst, whereas in LiDAR and radar it outputs an empty point cloud or an all-zero frame; the frame-repeating faults of these two modalities are the microelectronics freeze (LiDAR) and the inference timing fault (radar), which repeat the last healthy frame preceding the burst. In all three modalities, a temporal misalignment delivers a frame delayed by 1, 2, 4, or 8 frames depending on severity, corresponding to 0.1--0.8~s at the 10~Hz of the camera and the LiDAR and 0.25--2.0~s at the 4~Hz of the radar. Two synergistic routines build on the delayed frame: for cameras, temporal ghosting superimposes it on the current frame, which reproduces the duplicated appearance of moving objects and is the camera-only counterpart of the temporal misalignment and multipath interaction; for radar, the temporal-multipath synergy injects multipath ghost echoes onto the delayed frame.

Each run produces the corrupted stream together with a per-frame manifest that records the fault label, taxonomy key, category, affected subsystem, severity, burst identifier, source frame for temporal faults, and the random seeds used (one per burst for fixed-pattern components and one per frame for frame-varying ones), as well as the exact injection plan. Every corrupted frame can thus be regenerated from the clean frame and the recorded seeds, and a whole run is reproduced by repeating it with the same master seed.

\subsection{Illustrative Examples}
\label{sec:examples}
Each example applies one fault per taxonomy category (hardware, environmental, calibration, software, and synergistic) at severity level \textit{severe} to a single frame per modality, namely the first camera, LiDAR, and radar frames of the Boreas sequence \texttt{boreas-2020-11-26-13-58}. The remaining routines, as well as the sequence-level faults, which depend on the temporal context, are illustrated in the supplementary material.

\subsubsection{Camera}
\begin{figure*}[!t]
\centering
\includegraphics[width=\textwidth]{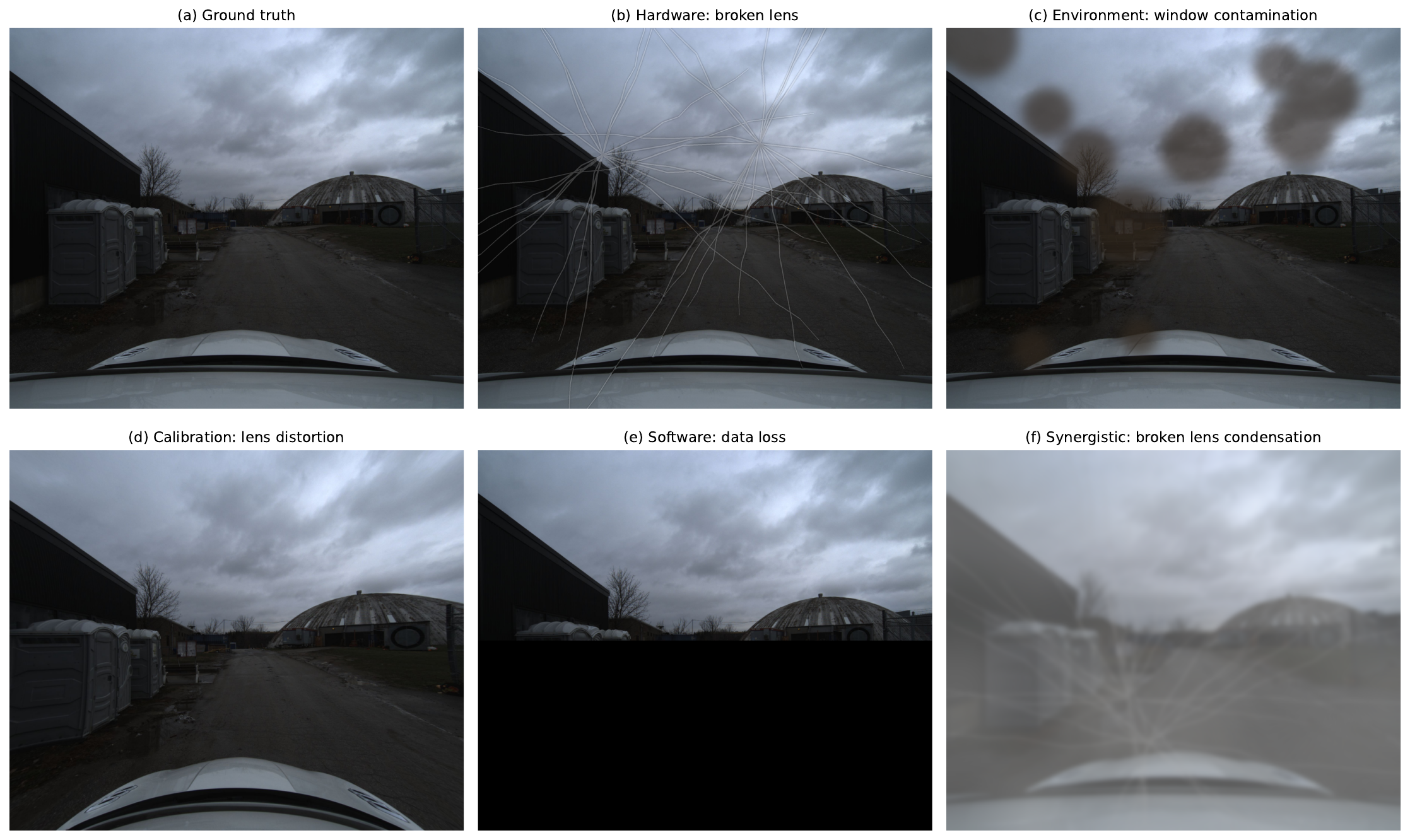}
\caption{Examples of camera fault injection on a frame of the Boreas dataset, one fault per taxonomy category at severity level \textit{severe}. (a)~Ground truth. (b)~Hardware: broken lens, with radial cracks and local refraction. (c)~Environmental: window contamination, with semi-opaque deposits on the protective window. (d)~Calibration: residual barrel lens distortion. (e)~Software: data loss, with the lower half of the frame not transmitted. (f)~Synergistic: broken lens combined with condensation.}
\label{fig:camera_faults_simulation}
\end{figure*}

The synthetic injection processes replicate the observable symptoms identified in the taxonomy. Fig.~\ref{fig:camera_faults_simulation} shows one fault per category applied to the camera frame: a broken lens (hardware), window contamination (environmental), residual lens distortion (calibration), data loss with a truncated frame (software), and a broken lens combined with condensation (synergistic). The remaining camera routines are illustrated in Section~S1 of the supplementary material.

\subsubsection{LiDAR}
\begin{figure*}[!t]
\centering
\includegraphics[width=\textwidth]{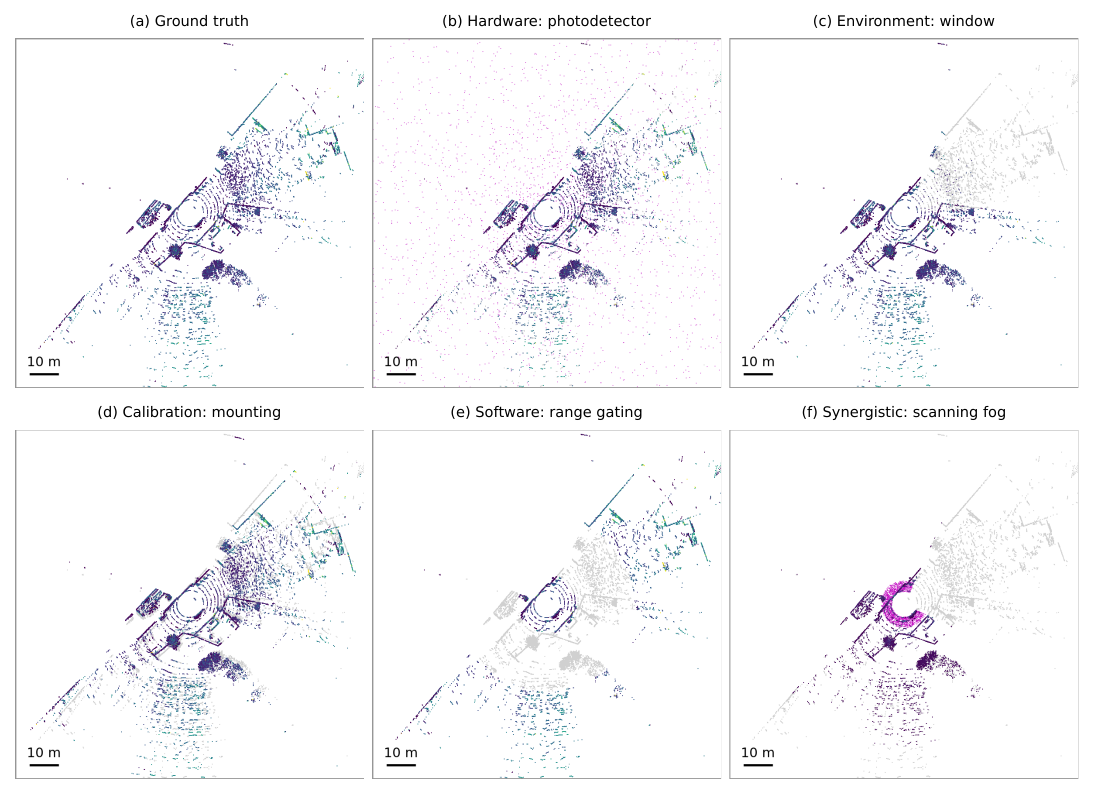}
\caption{Examples of LiDAR fault injection on a frame of the Boreas dataset, one fault per taxonomy category at severity level \textit{severe}, in a top-down view ($\pm 60$~m). Points are colored by intensity (square-root scaling); light-gray points are returns of the clean frame that are missing or displaced, and magenta points are spurious returns introduced by the fault. (a)~Ground truth. (b)~Hardware: photodetector fault, with loss of returns and spurious noise points. (c)~Environmental: window contamination, with a blind patch on the protective window. (d)~Calibration: extrinsic mounting misalignment. (e)~Software: range gating, with a band of ranges discarded by the filter. (f)~Synergistic: scanning fault combined with fog, with a lost angular sector and near-field backscatter.}
\label{fig:lidar_faults_simulation}
\end{figure*}

Fig.~\ref{fig:lidar_faults_simulation} applies the same selection principle to the LiDAR frame: a photodetector fault (hardware), window contamination (environmental), extrinsic mounting misalignment (calibration), range gating (software), and a scanning fault combined with fog (synergistic). The LiDAR routines operate on the point cloud in the spherical geometry of the sensor and act by removing returns, adding spurious points, attenuating intensities, or applying rigid transformations; spurious points are flagged by undefined ring and time fields so that they remain identifiable. Each panel is a top-down view colored by return intensity, with the clean frame drawn in light gray behind the faulty one so that removed and displaced returns remain visible, and with spurious returns in magenta. For the three faults whose visible effect depends on where the random pattern falls (scanning, window, and scanning combined with fog), the seed was selected among 100 candidates to maximize the number of hidden distant returns, for illustration purposes only; the selected seeds are recorded in the figure manifest of the toolkit. The remaining lidar routines are illustrated in Section~S2 of the supplementary material.

\subsubsection{Radar}
\begin{figure*}[!t]
\centering
\includegraphics[width=\textwidth]{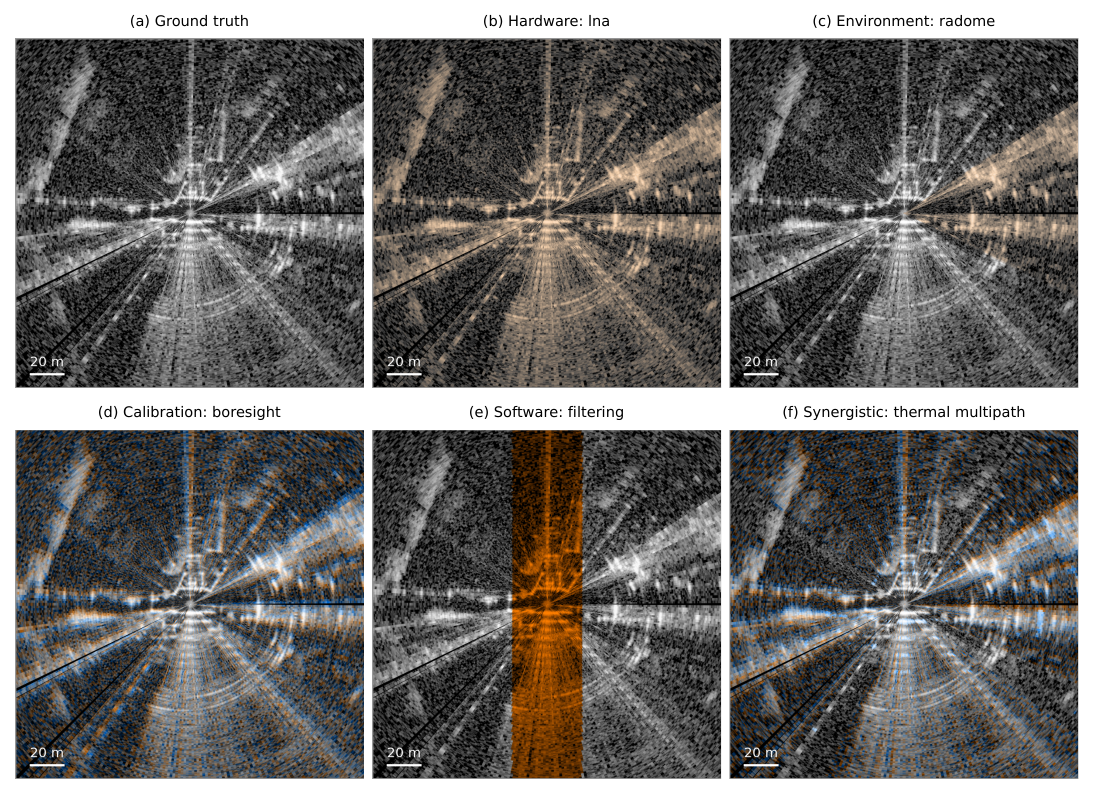}
\caption{Examples of radar fault injection on a frame of the Boreas dataset, one fault per taxonomy category at severity level \textit{severe}, rendered in a top-down Cartesian view with echo power on a square-root scale. In the faulty panels, gray indicates unchanged returns, orange indicates attenuated or discarded returns, and blue indicates added noise, ghost echoes, or displaced echoes. (a)~Ground truth (grayscale). (b)~Hardware: LNA degradation, causing uniform receiver gain loss across echoes. (c)~Environmental: radome contamination, attenuating returns within a localized azimuth sector. (d)~Calibration: boresight misalignment, rotating echoes in azimuth. (e)~Software: ROI filtering, blanking vertical Cartesian bands. (f)~Synergistic: thermal packaging deformation combined with multipath ghost reflections.}
\label{fig:radar_faults_simulation}
\end{figure*}

Fig.~\ref{fig:radar_faults_simulation} applies the same selection principle to the radar frame: LNA degradation (hardware), radome contamination (environmental), boresight misalignment (calibration), region-of-interest filtering (software), and thermal packaging deformation combined with multipath (synergistic). The radar routines operate directly on the 2D polar power spectrum (azimuth rows by range bins) while preserving the frame header columns, and act by scaling echo power above the estimated noise floor, injecting exponential clutter or interference noise, shifting or smoothing azimuth rows, or blanking spatial bands. Each panel is rendered in a top-down Cartesian projection on a black background, with echo power shown on a square-root scale. The ground-truth frame is displayed in grayscale, whereas the faulty panels use a gray--orange--blue composite in which gray denotes unchanged echoes, orange denotes attenuated or discarded returns, and blue denotes added noise, ghost echoes, or angularly displaced returns. For the two faults whose visible effect depends on where the random pattern falls (radome contamination and thermal deformation combined with multipath), the seed was selected among 100 candidates to maximize the difference from the clean frame, for illustration purposes only. The remaining radar routines are illustrated in Section~S3 of the supplementary material.

\subsection{Operational Scope and Design Choices}
\label{sec:scope}

To be portable across public autonomous driving datasets without requiring proprietary sensor firmware, specialized hardware-in-the-loop simulators, or raw ADC waveforms, the toolkit operates directly at the primary data representation level typically used by perception and diagnostic stacks: RGB images, 3D point clouds, and 2D polar radar power spectra. This design choice establishes a practical balance between phenomenological fidelity and cross-dataset portability. For cameras, injecting anomalies at the post-ISP representation allows the framework to emulate both optical disturbances and the symptoms of embedded processing faults across standard vision pipelines without relying on sensor-specific RAW Bayer formats. For LiDAR, converting point coordinates, intensities, and metadata into spherical coordinates enables plausible geometric corruptions while maintaining direct compatibility with standard 3D perception backbones; points synthesized by a fault routine carry \texttt{NaN} in their \texttt{ring} and \texttt{time} fields so that they remain programmatically identifiable. Similarly, targeting the 2D polar range-azimuth power spectrum of the Navtech CIR304-H directly reflects how scanning radar data are structured in benchmark datasets such as Boreas. The four graded severity levels (\textit{mild} to \textit{extreme}) introduce increasing, reproducible degradation patterns suitable for sensitivity benchmarking and for stress-testing perception and fault diagnosis algorithms on time-stamped, real-world multi-modal recordings with ground-truth fault annotations. However, the routines are functional and phenomenological proxies of observable symptoms rather than physically based simulations of the underlying wave propagation. 

Finally, the illustrative examples presented in Section~\ref{sec:examples} are qualitative and evaluate one frame per modality. Severity presets across all modalities are not calibrated against physical measurements of hardware-degraded sensors; their empirical calibration, alongside a quantitative comparison with real adverse-weather recordings (such as the rain and snow sequences available in Boreas), is left to future work. The toolkit does not replace full-wave electromagnetic, radiative-transfer, or physics-based optical simulation, but provides a controlled, reproducible testbed for robust perception and fault diagnosis research.

\section{Discussion: Implications for Functional Safety and FDD}
\label{sec:discussion}

Multi-sensor perception pipelines rely on redundancy across camera, LiDAR, and radar to maintain environmental awareness under degraded sensing conditions. The taxonomy suggests, however, that redundancy alone may not ensure dependable operation when degradations are correlated across modalities, originate upstream of the fusion, or affect the sensor itself, and that dedicated Fault Detection and Diagnosis (FDD) is needed for perception tasks such as object detection, tracking, segmentation, and mapping.

\subsection{Limits of Redundancy and Synergistic Faults}
Fusion algorithms often implicitly assume that degradations are independent across modalities and that the measurements are mutually consistent. This holds reasonably well for isolated environmental conditions, such as a camera blinded by direct sunlight while a radar keeps its range measurements, but it can break down for three categories. Software faults (Category~IV) corrupt or delay the data before fusion takes place, for instance through frame freezes or latency that compromise temporal fusion. Calibration faults (Category~III) affect the sensor transformations globally and produce spatial offsets that adaptive feature weighting cannot resolve without active calibration monitoring, while each sensor still appears healthy. Synergistic faults (Category~V) couple external disturbances with internal wear or spatial drift; because their symptoms may mimic natural scene variability or dense clutter, passive fusion may incorporate them as false detections or misaligned objects, and detectors designed for single faults may miss them.

\subsection{Traceability to ISO~26262 and ISO~21448 (SOTIF)}
The categories of the taxonomy map indicatively onto the two standards:
\begin{itemize}
    \item \textbf{ISO~26262} addresses hazards caused by the malfunctioning behavior of safety-related electrical and electronic systems\cite{ISO26262_1}. Hardware (I), calibration (III), and software (IV) faults fall mainly within its scope: hardware degradations, such as LNA gain loss or photodetector sensitivity loss, correspond to random hardware failures, whereas firmware crashes, packet misordering, and ISP failures correspond to systematic faults of the sensor logic.
    \item \textbf{ISO~21448 (SOTIF)} addresses hazards due to functional insufficiencies, including performance insufficiencies of electrical and electronic elements, and does not apply to faults covered by ISO~26262~\cite{ISO21448_2022}. Environmental faults (II), such as aerosol scattering, radome contamination, or multipath reflections, correspond to performance limitations that external triggering conditions bring to the surface.
\end{itemize}
Synergistic faults (V) sit at the intersection of the two, for instance when a lens fracture (hardware) enables moisture condensation (environmental). The assignment follows the root-cause criterion of Section~\ref{sec:taxonomy}; some entries, such as miscalibrations due to aging, may fall under either standard depending on the safety analysis.

\subsection{Testbed for FDD and Safety Analyses}
A perception stack that has to detect, isolate, and mitigate these failure modes needs diagnosis that goes beyond binary fault flags and identifies the root-cause category and the affected subsystem. The categories suggest different detection strategies: calibration faults call for cross-modal consistency monitors~\cite{Peng2024Miscalibration,Tahiraj2025CalorNoCal}; hardware and environmental degradations can be addressed with per-modality health indicators and multi-sensor fault detection and diagnosis frameworks~\cite{MultiSensorFaultDetection,Baseline,Sensor_fault_diagnosis}; and software faults, which often produce missing, stale, or empty frames, call for data-integrity checks on timestamps and frame sequences. The temporal behavior of a fault (permanent, intermittent, or transient) additionally determines the observation window and the persistence logic of a detector. Once a fault is identified, the fusion can downweight or exclude the unreliable modality~\cite{MultiModalExpert}, in line with architectures that separate sensor health from task confidence~\cite{viktor2026modality}.

Developing such diagnosis has been hindered by the scarcity of public multi-modal datasets with labeled sensor anomalies beyond adverse weather~\cite{BoreasDataset,TWICEdataset,SIDdataset,DAWNdataset,WADSdataset}. The toolkit of Section~\ref{sec:toolkit} addresses this gap with 94 symptom-level routines across camera, LiDAR, and radar data, plausible fault bursts with controlled temporal behavior and graded severity, and per-frame manifests recording the fault category, affected subsystem, severity, and source frame. These annotations make it possible to benchmark analytical-redundancy classifiers and learned anomaly detectors, to measure detection rate and latency per fault class, and to quantify the sensitivity of fusion stacks to single and compound faults. In safety analyses, where fault injection is a common means to demonstrate the diagnostic coverage of safety mechanisms\cite{ISO26262_1}, the toolkit complements hardware-level campaigns and real degraded recordings at the data level. Because the injection is phenomenological and the severity presets are not calibrated, results should be read as sensitivity evidence rather than as quantitative safety figures, and diagnosis trained on synthetic faults requires validation on real degraded data.

\section{Conclusion}
\label{sec:conclusion}

This paper reviewed the literature on exteroceptive sensor faults and introduced a cross-modal, origin-driven taxonomy of camera, LiDAR, and radar faults in autonomous vehicles. By categorizing anomalies into hardware, environmental, calibration, software, and synergistic classes, the taxonomy traces each degradation from its physical or logical root cause and affected subsystem to its observable manifestation in the sensor data, relates the categories to the analyses of ISO~26262 and ISO~21448 (SOTIF), and indicates why redundancy alone may be insufficient for software, calibration, and synergistic faults. To operationalize the taxonomy, we released an open-source fault injection toolkit with 94 symptom-level routines, graded severity levels, burst scheduling, and per-frame ground-truth annotations, whose outputs we illustrated qualitatively on the Boreas dataset.

Since the routines are phenomenological proxies and their severity presets are not calibrated, future work will focus on three directions: 1)~calibrating the presets against measurements of degraded sensors and real adverse-weather recordings; 2)~extending the radar routines to next-generation 4D imaging radars incorporating Doppler and elevation channels; and 3)~using the toolkit to quantitatively benchmark perception and fusion models and to develop and evaluate Fault Detection and Diagnosis architectures for fault-tolerant multi-sensor fusion.

%{\appendices
%\section*{Proof of the First Zonklar Equation}
%Appendix one text goes here.
% You can choose not to have a title for an appendix if you want by leaving the argument blank
%\section*{Proof of the Second Zonklar Equation}
%Appendix two text goes here.}

\bibliographystyle{IEEEtran}
\bibliography{references}

\begin{IEEEbiographynophoto}{Davide Possenti} is currently a Ph.D. Candidate in Mechanical Engineering at Politecnico di Milano, Milan, Italy. He received his B.S. degree in Civil Engineering and his M.S. degree in Mobility Engineering from Politecnico di Milano. His research interests include autonomous vehicles, with a particular emphasis on perception and localization under adverse weather conditions and sensor failures, multi-modal sensor fusion, and generative modeling for geometric scene representation.
\end{IEEEbiographynophoto}

\begin{IEEEbiographynophoto}{Daniele Cianca} received his B.S. degree in Mechanical Engineering and his M.S. degree in Mechanical Engineering, specializing in Mechatronics and Robotics, from Politecnico di Milano. His research interests include autonomous vehicles, with a particular emphasis on perception and localization under adverse weather conditions and sensor failures.
\end{IEEEbiographynophoto}

\begin{IEEEbiographynophoto}{Stefano Arrigoni} is currently an Assistant Professor with the Department of Mechanical Engineering, Politecnico di Milano, Milan, Italy. He received the M.S. degree in mechanical engineering and the Ph.D. degree in applied mechanics from Politecnico di Milano in 2013 and 2017, respectively. His research interests include autonomous and connected vehicles, with a focus on motion planning and control, model predictive control, cooperative localization and perception, and V2X-enabled ADAS.
\end{IEEEbiographynophoto}

\begin{IEEEbiographynophoto}{Francesco Braghin} received the M.S. degree and the Ph.D. degree in applied mechanics from Politecnico di Milano, Milan, Italy, in 1997 and 2001, respectively. He became Associate Professor in 2011 and Full Professor of Applied Mechanics in 2015. His research interests include vehicle dynamics and control, autonomous and connected vehicle systems, mechatronics, robotics, and vibration-control and metamaterial-based systems.
\end{IEEEbiographynophoto}

\vfill
\clearpage
\onecolumn % Forza il layout a singola colonna da qui in poi

% Reset dei contatori per avere una numerazione indipendente (es. Fig S1)
\setcounter{page}{1}
\setcounter{section}{0}
\setcounter{figure}{0}
\setcounter{table}{0}
\renewcommand{\thefigure}{S\arabic{figure}}
\renewcommand{\thetable}{S\arabic{table}}
\renewcommand{\thepage}{S\arabic{page}}

\begin{center}
    {\huge \bfseries Supplementary Material for:\\[1ex]
    Exteroceptive Sensor Faults in Autonomous Vehicles:\\
    A Cross-Modal Taxonomy and Fault Injection Toolkit\par}
    
    \vspace{0.8cm}
    
    {\large Davide Possenti, Daniele Cianca, Stefano Arrigoni, Francesco Braghin\par}
\end{center}
\vspace{1cm}

\input{supplementary.tex}

\input{supplementary.tex}
\end{document}

%% file: supplementary.tex
\section{Camera Fault Catalog}
\label{supp:fault_catalog}

\begin{figure}[!t]
\centering
\includegraphics[width=\textwidth]{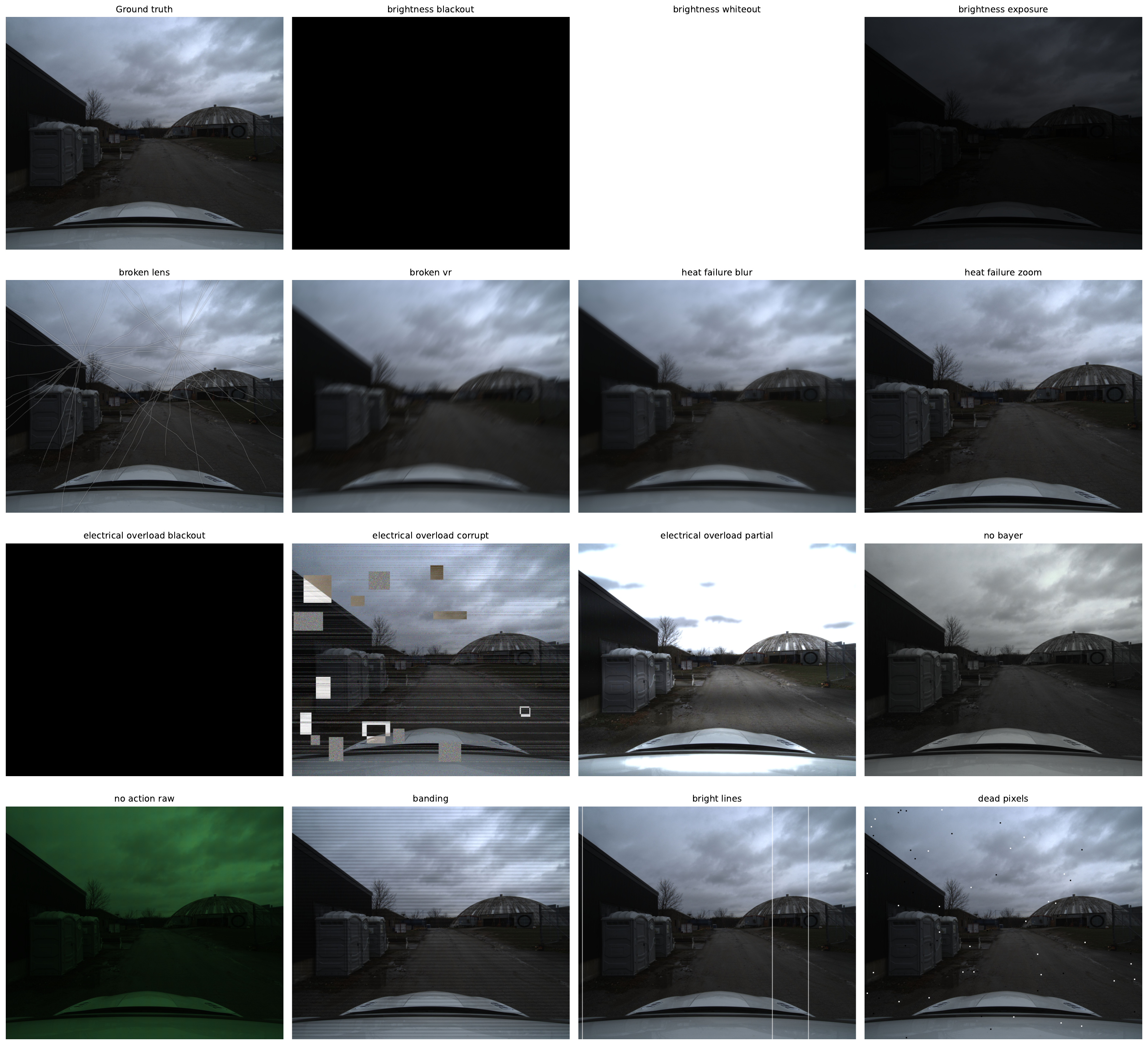}
\caption{Hardware faults. After the ground truth, from left to right and top to bottom: brightness blackout, whiteout, and exposure; broken lens; broken vibration reduction (motion blur); heat failure (blur and zoom); electrical overload (blackout, corruption, and partial); absent Bayer filter; ISP raw output (no action); readout banding; bright lines; and dead pixels.}
\label{fig:catalog_hardware}
\end{figure}

\begin{figure}[!h]
\centering
\includegraphics[width=\textwidth]{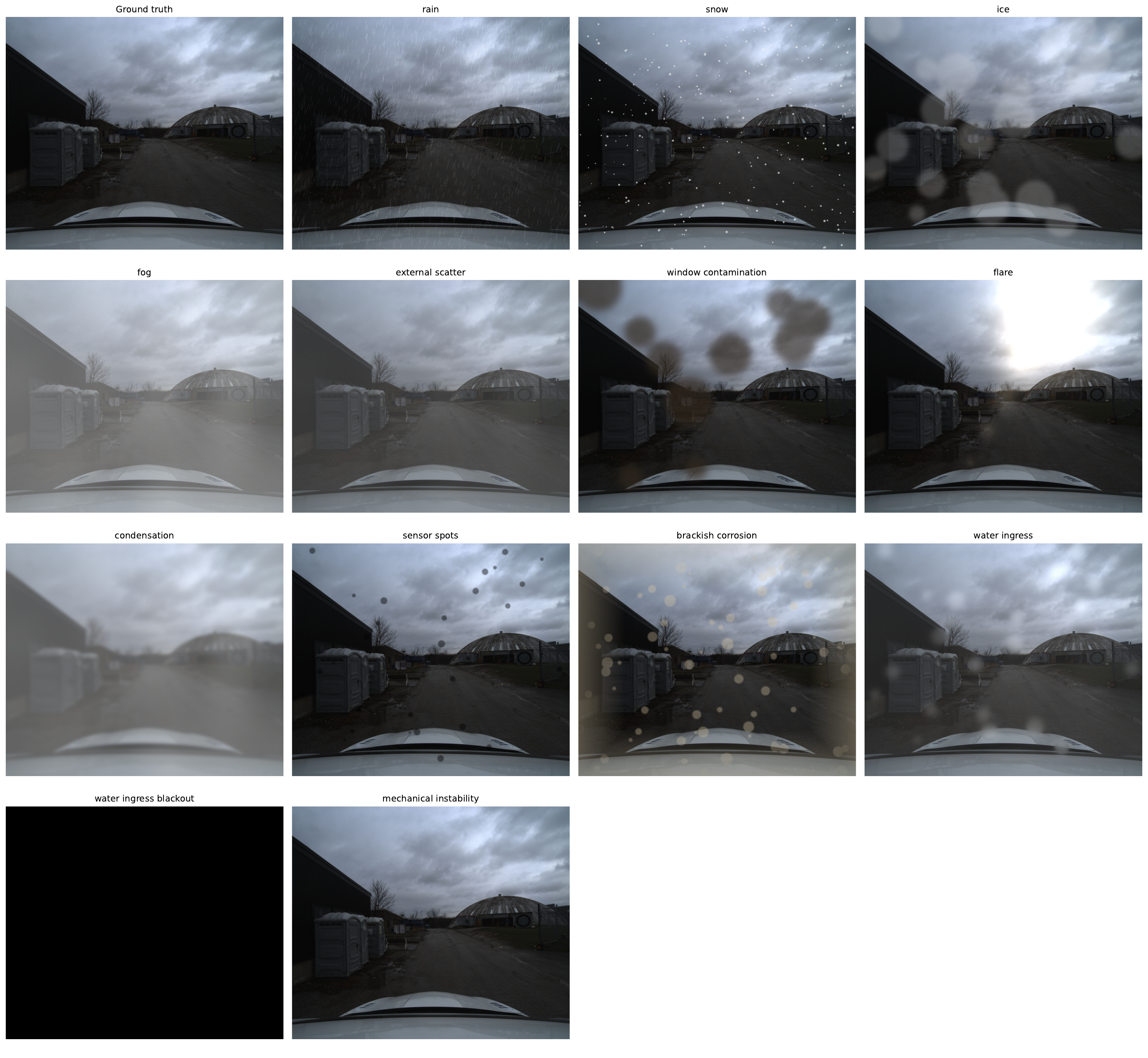}
\caption{Environmental faults. After the ground truth: rain, snow, ice, fog, external scatter, window contamination, flare, condensation, sensor spots, brackish corrosion, water ingress, water ingress blackout, and mechanical instability.}
\label{fig:catalog_environment}
\end{figure}

\begin{figure}[!h]
\centering
\includegraphics[width=\textwidth]{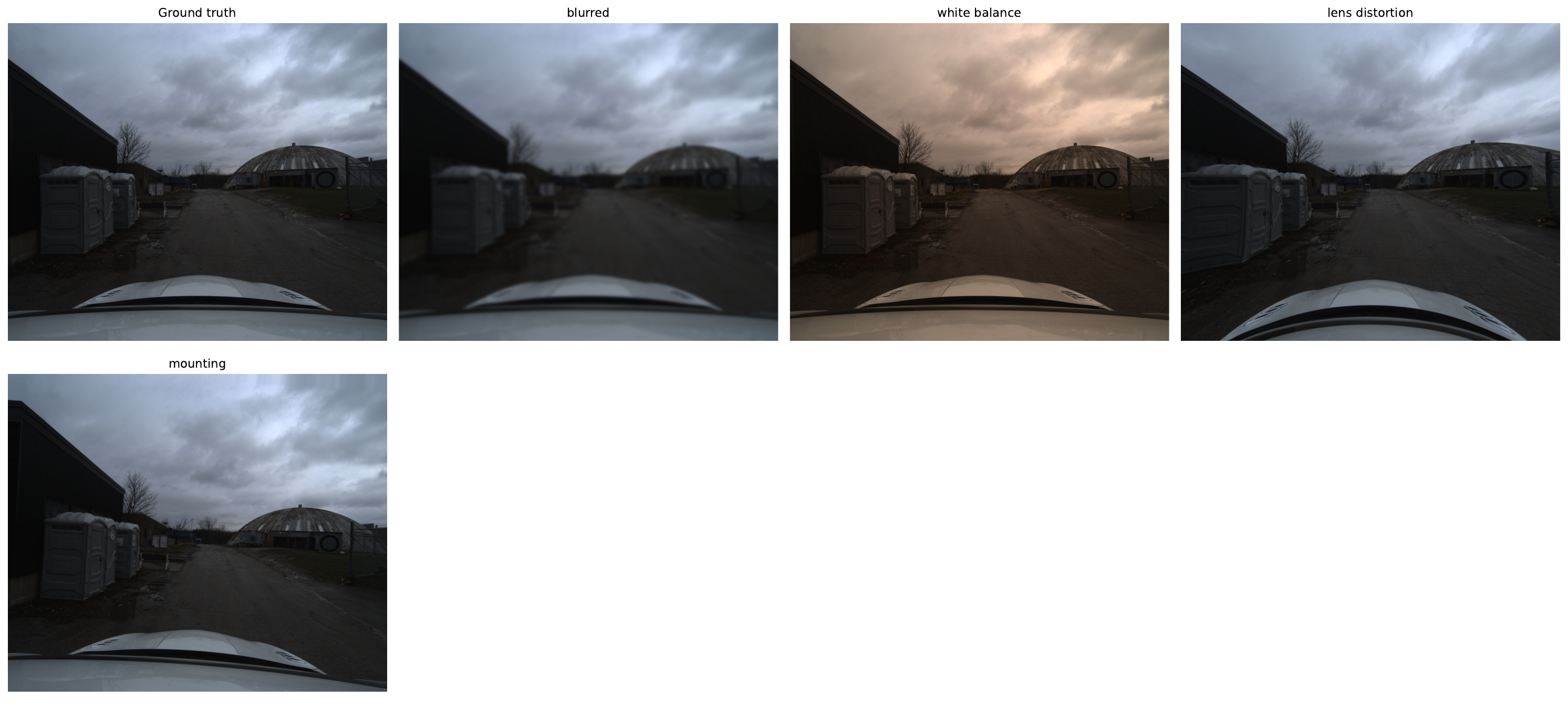}
\caption{Calibration faults. After the ground truth: blurred image, white balance error, lens distortion error, and extrinsic misalignment (mounting).}
\label{fig:catalog_calibration}
\end{figure}

\begin{figure}[!h]
\centering
\includegraphics[width=\textwidth]{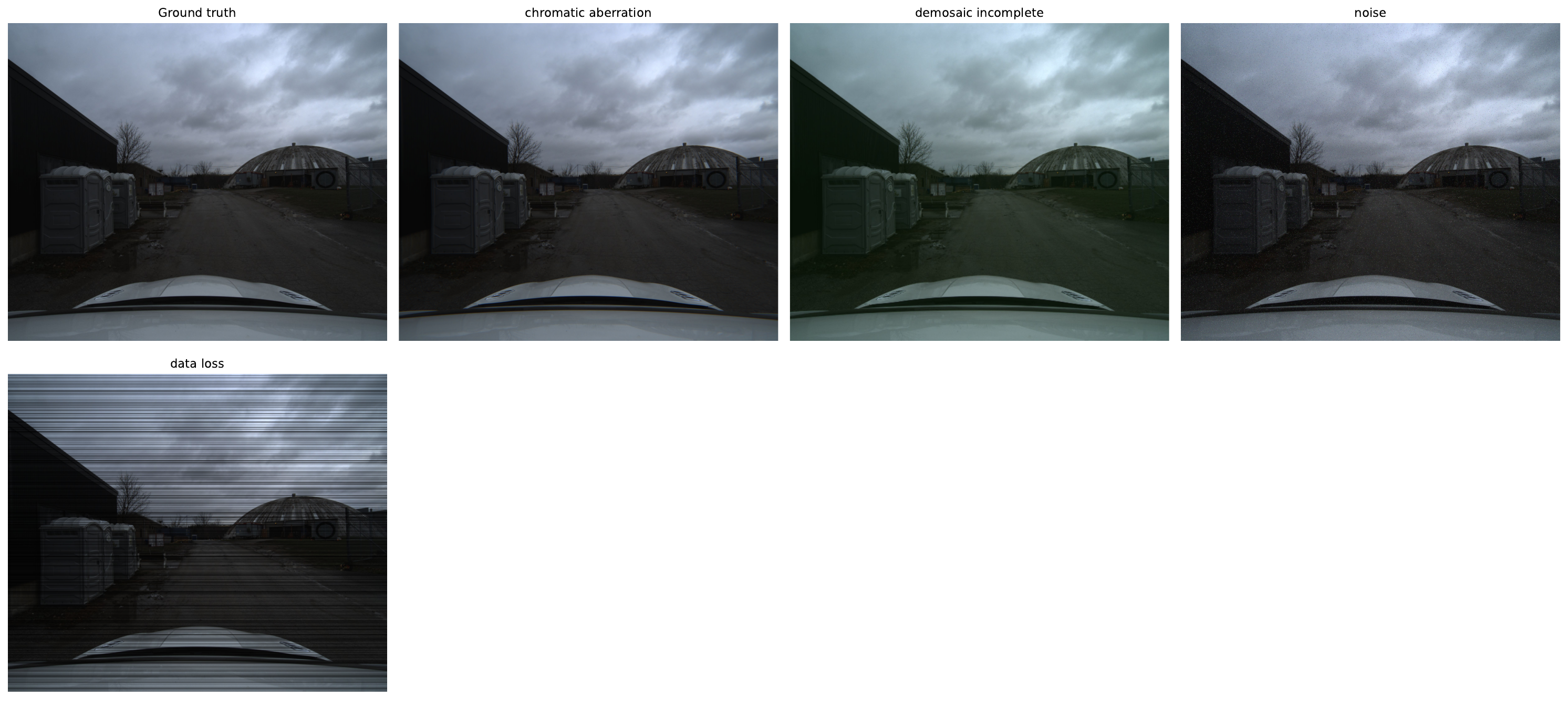}
\caption{Software faults. After the ground truth: chromatic aberration, incomplete demosaicing, incomplete noise reduction (noise), and data loss (shown in its row-loss mode).}
\label{fig:catalog_software}
\end{figure}

\begin{figure}[!h]
\centering
\includegraphics[width=\textwidth]{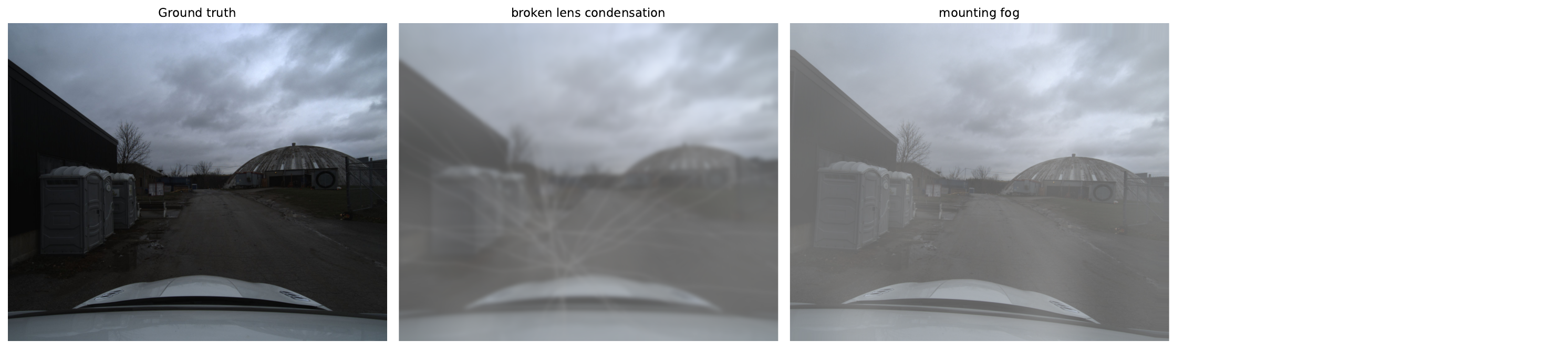}
\caption{Synergistic faults. After the ground truth: broken lens combined with condensation, and extrinsic misalignment (mounting) combined with fog.}
\label{fig:catalog_synergistic}
\end{figure}

This section illustrates the camera injection routines of the toolkit that are not shown in Fig.~4 of the main manuscript. % Hard-coded reference to the main manuscript: re-check the figure number before submission.
All examples are applied to the same Boreas frame used in the main text, at severity level \textit{severe} and with fixed random seeds, so that the panels are directly comparable. Each panel is titled with the name of the corresponding routine. Figs.~\ref{fig:catalog_hardware}--\ref{fig:catalog_synergistic} cover the hardware (15 routines), environmental (13), calibration (4), software (4 of 6), and synergistic (2 of 3) categories. The remaining routines depend on the temporal context and are described in Section~\ref{supp:temporal}.

\subsection{Sequence-Level Faults}
\label{supp:temporal}
Three camera routines cannot be illustrated on a single frame because they depend on the temporal context. \textit{Silent failure} emulates a frozen stream by repeating the last healthy frame preceding the burst, and has no severity scale. \textit{Temporal misalignment} delivers, for every frame of the burst, the frame acquired $d$ frames earlier. \textit{Temporal ghosting}, the camera-only synergistic counterpart of the temporal misalignment and multipath interaction, blends the current frame $I_t$ with the delayed frame $I_{t-d}$ as $\tilde{I}_t = (1-\alpha)\,I_t + \alpha\,I_{t-d}$. Table~\ref{tab:temporal_parameters} reports the delay and the ghost opacity for each severity level.

\begin{table}[!h]
\caption{Parameters of the Sequence-Level Camera Faults}
\label{tab:temporal_parameters}
\centering
\begin{tabular}{lcc}
\toprule
Severity & Delay $d$ (frames) & Ghost opacity $\alpha$ \\
\midrule
Mild     & 1 & 0.10 \\
Moderate & 2 & 0.20 \\
Severe   & 4 & 0.35 \\
Extreme  & 8 & 0.50 \\
\bottomrule
\end{tabular}
\end{table}

\subsection{Notes on the Examples}
Some effects are inherently weak at the scale of the figures. Chromatic aberration produces color fringes of a few pixels, confined to high-contrast edges near the image border, and incomplete demosaicing appears mainly as a color cast because its block structure is below the display resolution. Dead pixels are rendered as small clusters whose size scales with the image resolution so that they remain visible, whereas real defective pixels are typically isolated. Data loss is shown in its row-loss mode; the toolkit also supports block loss and truncated frames.

\clearpage
\section{LiDAR Fault Catalog}
\label{supp:lidar_catalog}

\begin{figure}[!t]
\centering
\includegraphics[width=\textwidth]{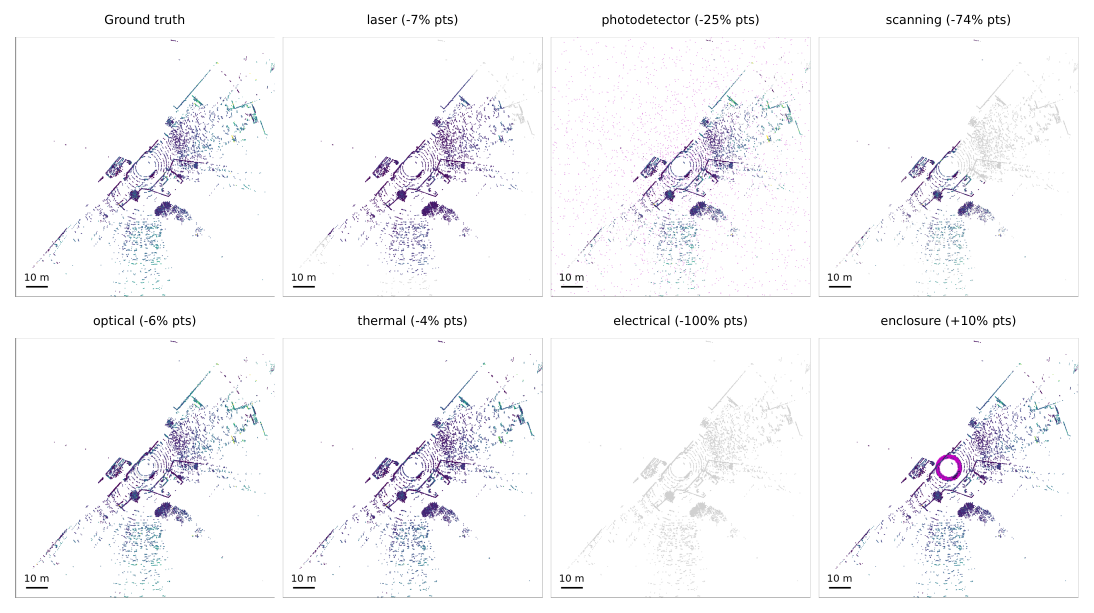}
\caption{LiDAR hardware faults. After the ground truth: laser device, photodetector, scanning mechanism, optical system, thermal management, electrical connection, and enclosure faults.}
\label{fig:catalog_lidar_hardware}
\end{figure}

\begin{figure}[!t]
\centering
\includegraphics[width=\textwidth]{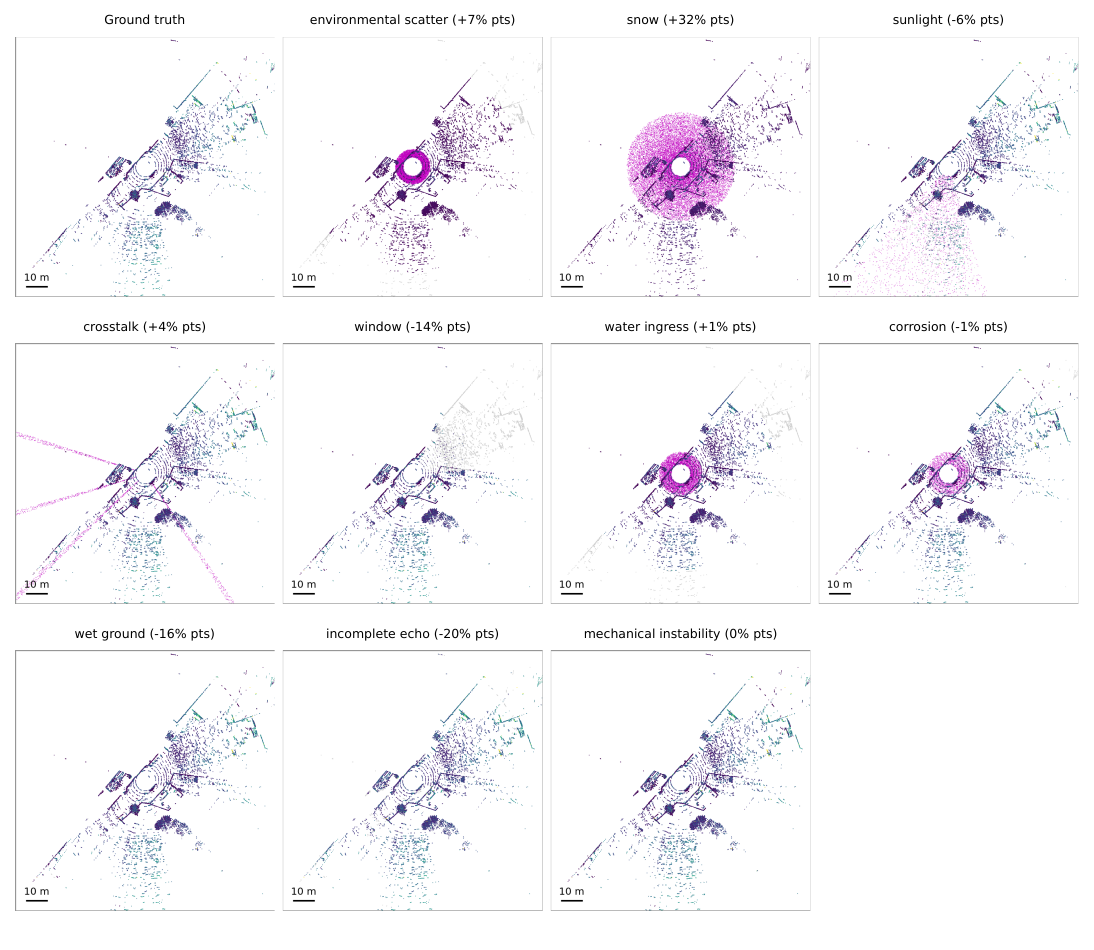}
\caption{LiDAR environmental faults. After the ground truth: atmospheric scattering (rain, fog, and external scatter), snow, strong sunlight, crosstalk, window contamination, water ingress, brackish corrosion, wet ground, incomplete echo, and mechanical instability.}
\label{fig:catalog_lidar_environment}
\end{figure}

\begin{figure}[!t]
\centering
\includegraphics[width=\textwidth]{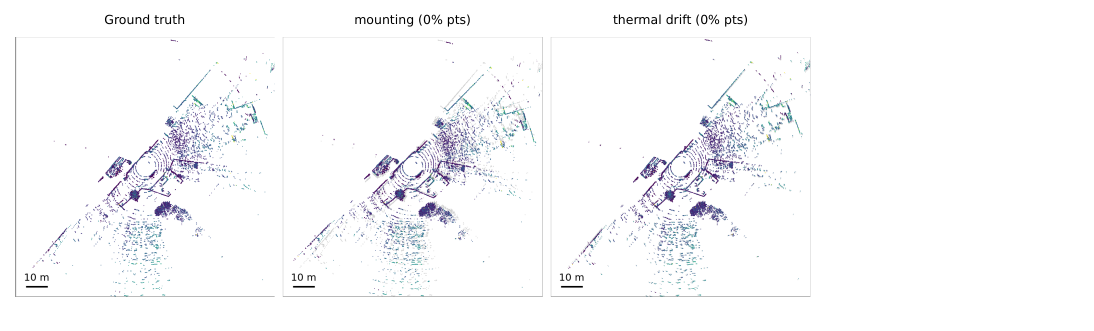}
\caption{LiDAR calibration faults. After the ground truth: extrinsic mounting misalignment and thermal expansion drift.}
\label{fig:catalog_lidar_calibration}
\end{figure}

\begin{figure}[!t]
\centering
\includegraphics[width=\textwidth]{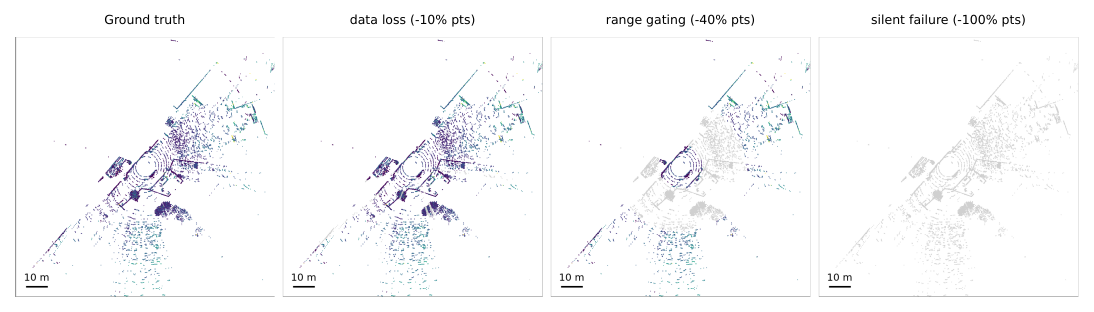}
\caption{LiDAR software faults. After the ground truth: data loss, range gating, and silent failure.}
\label{fig:catalog_lidar_software}
\end{figure}

\begin{figure}[!t]
\centering
\includegraphics[width=\textwidth]{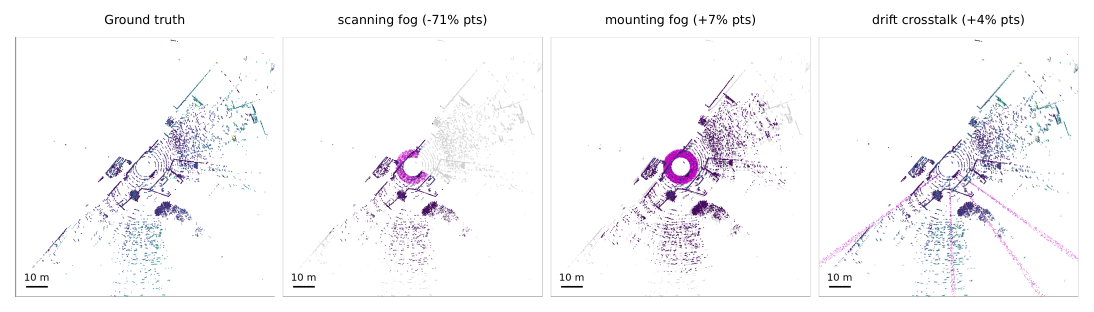}
\caption{LiDAR synergistic faults. After the ground truth: scanning fault combined with fog, extrinsic misalignment combined with fog, and calibration drift combined with crosstalk.}
\label{fig:catalog_lidar_synergistic}
\end{figure}

This section illustrates the LiDAR injection routines of the toolkit that are not shown in the main manuscript. All examples are applied to the same Boreas LiDAR frame used in the main text, at severity level \textit{severe} and with fixed random seeds. The representation matches the main text: points are colored by intensity (square-root scaling), light-gray points are returns of the clean frame that are missing or displaced, and magenta points are spurious returns. Each panel title reports the routine and the change in the number of points with respect to the clean frame. Figs.~\ref{fig:catalog_lidar_hardware}--\ref{fig:catalog_lidar_synergistic} cover the hardware (7 of 8 routines), environmental (10), calibration (2), software (3 of 4), and synergistic (3) categories. The two remaining routines depend on the temporal context and are described in Section~\ref{supp:lidar_temporal}.

\subsection{Sequence-Level Faults}
\label{supp:lidar_temporal}
Two LiDAR routines cannot be illustrated on a single frame because they depend on the temporal context. The \textit{microelectronics freeze} repeats the last healthy frame preceding the burst and has no severity scale. The \textit{temporal misalignment} delivers, for every frame of the burst, a frame acquired a severity-dependent number of frames earlier.

\subsection{Notes on the Examples}
Several routines modify the point cloud without removing returns, so their effect is not visible as a change in the number of points. The extrinsic mounting misalignment and the mechanical instability apply rigid rotations, which preserve the number of points and the measured ranges; the latter oscillates pitch and roll during the sweep and acts mainly on the elevation, which is barely visible in a top-down view. The thermal expansion drift is a range bias of 0.10~m at the severity level shown, which is below the resolution of the view. Faults that attenuate the intensity (e.g., laser, optical, and thermal faults) appear as a shift of the colors toward darker values. The snow routine models falling snow, with backscatter from flakes and attenuation, whereas deposits on the protective window are covered by the window contamination routine. The incomplete echo routine discards a fixed fraction of the weakest returns of each frame.

\clearpage
\section{Radar Fault Catalog}
\label{supp:radar_catalog}

\begin{figure}[!t]
\centering
\includegraphics[width=\textwidth]{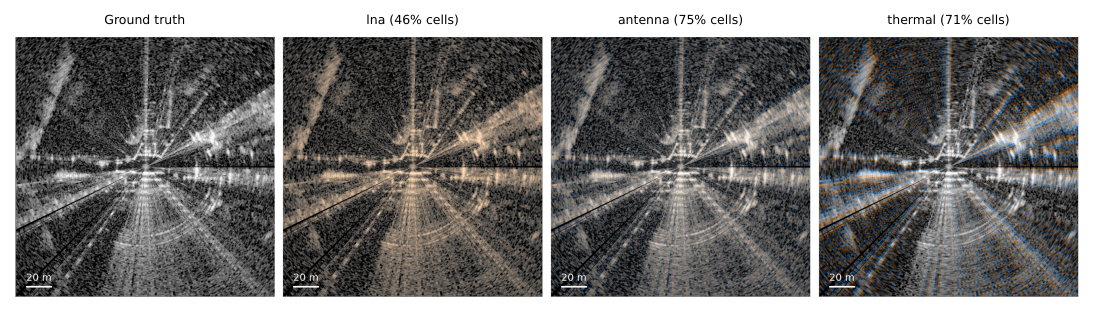}
\caption{Radar hardware faults. After the ground truth: LNA degradation, antenna array element failure, and thermal housing deformation.}
\label{fig:catalog_radar_hardware}
\end{figure}

\begin{figure}[!t]
\centering
\includegraphics[width=\textwidth]{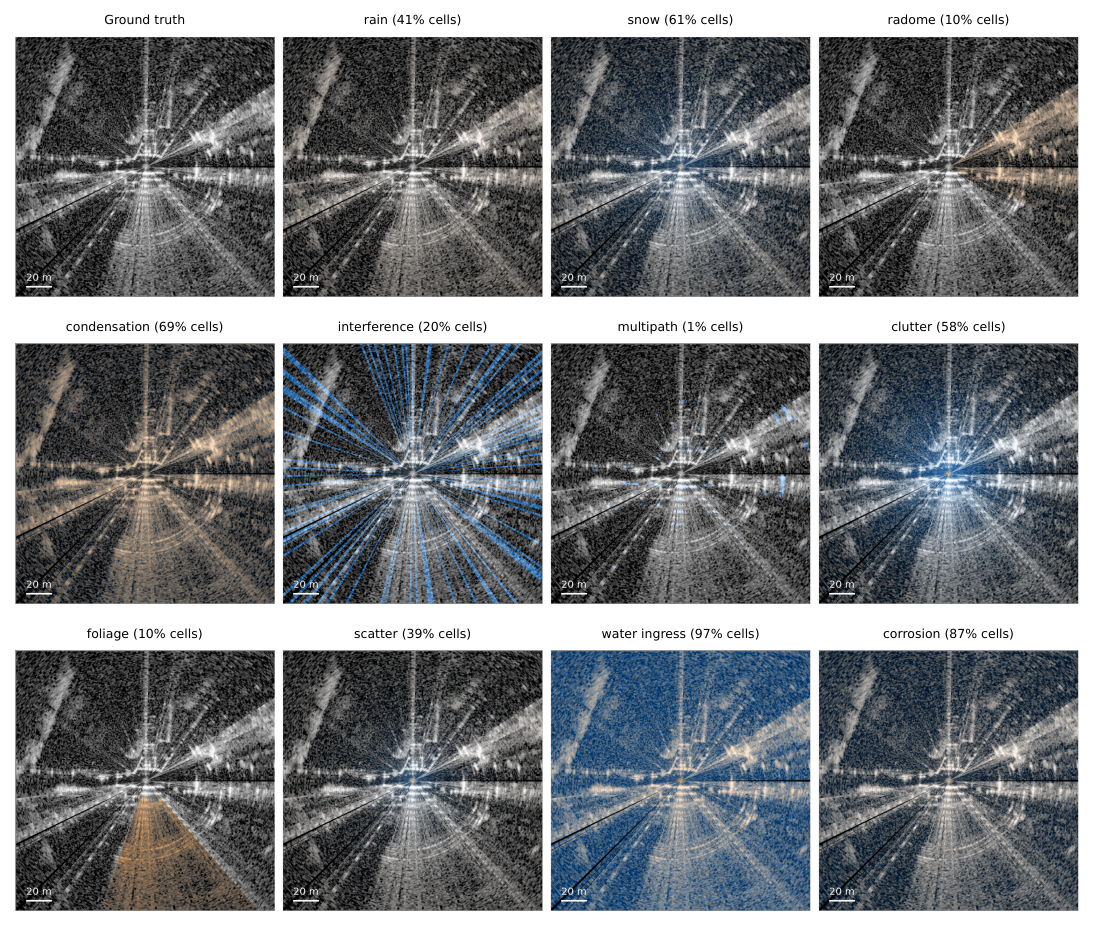}
\caption{Radar environmental faults. After the ground truth: rain attenuation, snow attenuation and backscatter, radome contamination, condensation film, radio interference, multipath ghosts, surface clutter, roadside foliage interference, external airborne scatter, water ingress, and brackish corrosion.}
\label{fig:catalog_radar_environment}
\end{figure}

\begin{figure}[!t]
\centering
\includegraphics[width=\textwidth]{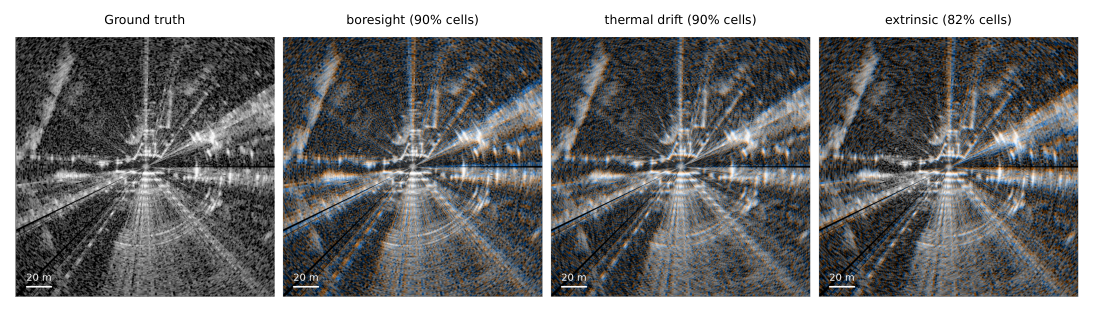}
\caption{Radar calibration faults. After the ground truth: boresight azimuth misalignment, thermal expansion drift, and extrinsic mounting misalignment.}
\label{fig:catalog_radar_calibration}
\end{figure}

\begin{figure}[!t]
\centering
\includegraphics[width=\textwidth]{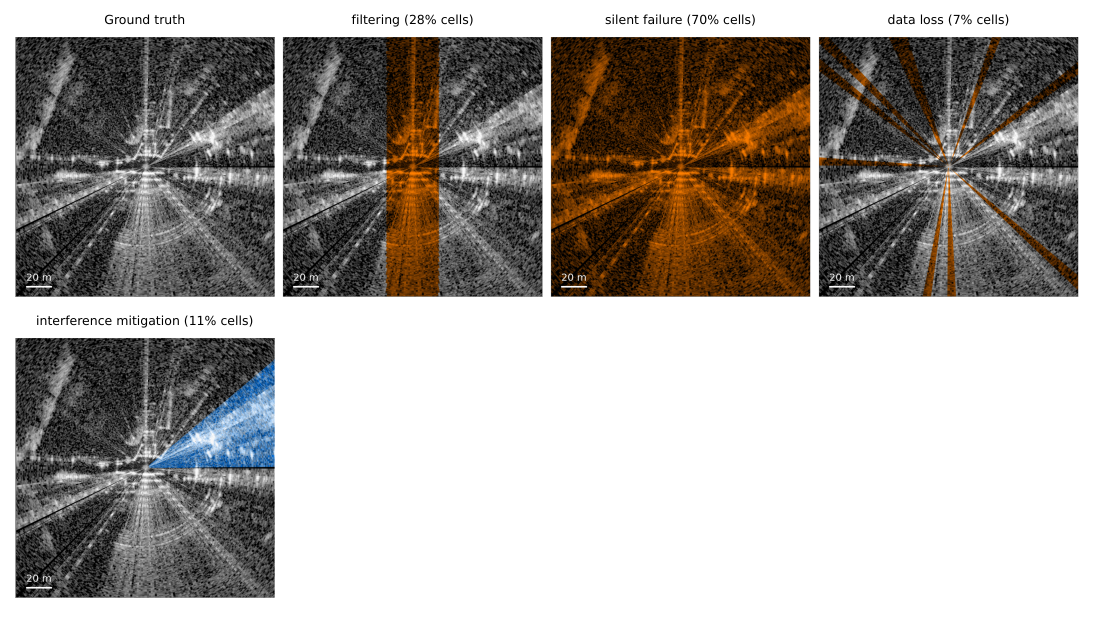}
\caption{Radar software faults. After the ground truth: ROI filtering, silent failure, packet data loss, and interference mitigation failure.}
\label{fig:catalog_radar_software}
\end{figure}

\begin{figure}[!t]
\centering
\includegraphics[width=\textwidth]{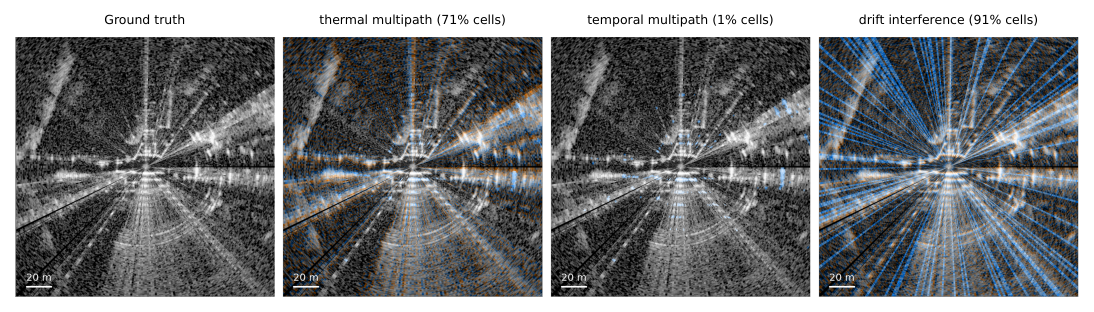}
\caption{Radar synergistic faults. After the ground truth: thermal deformation combined with multipath, temporal misalignment combined with multipath, and calibration thermal drift combined with radio interference.}
\label{fig:catalog_radar_synergistic}
\end{figure}

This section illustrates the radar injection routines of the toolkit that are not shown in the main manuscript. All examples are applied to the same Boreas radar frame used in the main text, at severity level \textit{severe} and with fixed random seeds, except for the radome contamination and the thermal deformation combined with multipath, whose seed is the one selected among 100 candidates. The visual representation matches the main text: the ground truth is rendered in grayscale, whereas faulty frames use a gray--orange--blue composite in which gray denotes unchanged power, orange denotes lost or attenuated power, and blue denotes added or displaced power. Each panel title reports the routine name and the percentage of polar cells whose 8-bit value changes with respect to the clean frame. Figs.~\ref{fig:catalog_radar_hardware}--\ref{fig:catalog_radar_synergistic} cover the hardware (3 routines), environmental (11), calibration (3), software (4 of 6), and synergistic (3) categories. The two remaining software routines depend on the temporal context and are described in Section~\ref{supp:radar_temporal}.

\subsection{Sequence-Level Faults}
\label{supp:radar_temporal}
Two radar software routines cannot be illustrated on a single static frame because they depend on the temporal sequence. \textit{Inference timing} models on-board scheduling delays by repeating the last healthy frame preceding the burst, and has no severity scale. \textit{Temporal misalignment} delivers, for every frame of the burst, the clean frame acquired $d \in \{1, 2, 4, 8\}$ frames earlier (corresponding to delays from 0.25~s to 2.0~s at the 4~Hz operating frequency of the Navtech radar). In the \textit{temporal multipath} synergy of Fig.~\ref{fig:catalog_radar_synergistic}, the multipath ghost model is applied to this delayed frame.

\subsection{Notes on the Examples}
For the frame shown, the multipath and temporal-multipath routines alter only about 1\% of the polar cells: only echoes exceeding the estimated noise floor by at least $6\sigma_n$, where $\sigma_n$ is the noise standard deviation, generate secondary double-bounce reflections at twice their true range, which appear as localized blue ghost arcs behind the strongest reflectors. The temporal-multipath panel of Fig.~\ref{fig:catalog_radar_synergistic} shows only this multipath component applied to the current frame, because the delay requires a frame sequence. Other localized faults, such as radome contamination (10\%) and packet data loss (7\%), also alter a small share of the cells. Conversely, angular displacements such as boresight misalignment (90\%), thermal deformation (71\%), and extrinsic misalignment (82\%) shift entire azimuth rows, producing paired orange--blue fringes along the edges of static structures. Water ingress (97\%) and brackish corrosion (87\%) act as system-level proxies that simultaneously attenuate target echoes and raise the global receiver noise floor, resulting in a widespread blue cast across the field of view. Finally, the silent failure outputs an all-zero frame, so the share of changed cells (about 70\%) equals the share of non-zero cells of the clean frame.

%% file: references.bib
@standard{ISO21448_2022,
  title        = {ISO 21448: Road Vehicles -- Safety of the Intended Functionality (SOTIF)},
  organization = {International Organization for Standardization},
  address      = {Geneva, Switzerland},
  year         = {2022},
  url          = {https://www.iso.org/standard/77490.html},
  note         = {First Edition, replaces ISO/PAS 21448:2019}
}

@inproceedings{Xia2025Misalignment,
  title={Robust long-range perception against sensor misalignment in autonomous vehicles},
  author={Xia, Zi-Xiang and Fadadu, Sudeep and Shi, Yi and Foucard, Louis},
  booktitle={2025 IEEE/CVF Winter Conference on Applications of Computer Vision (WACV)},
  pages={5761--5770},
  year={2025},
  organization={IEEE}
}

@article{Peng2024Miscalibration,
  title={Automatic miscalibration detection and correction of LiDAR and camera using motion cues},
  author={Peng, Pai and Pi, Dawei and Yin, Guodong and Wang, Yan and Xu, Liwei and Feng, Jiwei},
  journal={Chinese Journal of Mechanical Engineering},
  volume={37},
  number={1},
  pages={50},
  year={2024},
  publisher={Springer}
}

@inproceedings{Tahiraj2025CalorNoCal,
  title={Cal or No Cal?-Real-Time Miscalibration Detection of LiDAR and Camera Sensors},
  author={Tahiraj, Ilir and Swadiryus, Jeremialie and Fent, Felix and Lienkamp, Markus},
  booktitle={2025 IEEE/RSJ International Conference on Intelligent Robots and Systems (IROS)},
  pages={16098--16105},
  year={2025},
  organization={IEEE}
}

@article{FADE2025,
  title={Testing the fault-tolerance of multi-sensor fusion perception in autonomous driving systems},
  author={Tian, Haoxiang and Ding, Wenqiang and Han, Xingshuo and Wu, Guoquan and Guo, An and Zhang, Junqi and Chen, Wei and Wei, Jun and Zhang, Tianwei},
  journal={Proceedings of the ACM on Software Engineering},
  volume={2},
  number={ISSTA},
  pages={777--800},
  year={2025},
  publisher={ACM, New York, NY, USA}
}

@inproceedings{OnlineImageFaults,
  title={Online image sensor fault detection for autonomous vehicles},
  author={Chen, Yizhi and Zhu, Wenyao and Chen, Dejiu and Lu, Zhonghai},
  booktitle={2022 IEEE 15th International Symposium on Embedded Multicore/Many-core Systems-on-Chip (MCSoC)},
  pages={120--127},
  year={2022},
  organization={IEEE}
}

@article{SynthNoise2025,
  title={Synthesizing and identifying noise levels in autonomous vehicle camera radar datasets},
  author={Morales, Mathis and Habibi, Golnaz},
  journal={arXiv preprint arXiv:2505.00584},
  year={2025}
}

@inproceedings{FastAnomalyImage,
  title={Fast Image-Anomaly Mitigation for Autonomous Mobile Robots},
  author={Fumagalli, Gianmario and Huber, Yannick and Dymczyk, Marcin and Siegwart, Roland and Dub{\'e}, Renaud},
  booktitle={2021 IEEE/RSJ International Conference on Intelligent Robots and Systems (IROS)},
  pages={2436--2442},
  year={2021},
  organization={IEEE}
}

@article{BoreasDataset,
  title={Boreas: A multi-season autonomous driving dataset},
  author={Burnett, Keenan and Yoon, David J and Wu, Yuchen and Li, Andrew Z and Zhang, Haowei and Lu, Shichen and Qian, Jingxing and Tseng, Wei-Kang and Lambert, Andrew and Leung, Keith YK and others},
  journal={The International Journal of Robotics Research},
  volume={42},
  number={1-2},
  pages={33--42},
  year={2023},
  publisher={SAGE Publications, Sage UK: London, England}
}

@article{DAWNdataset,
  title={Dawn: vehicle detection in adverse weather nature dataset},
  author={Kenk, Mourad A and Hassaballah, Mahmoud},
  journal={arXiv preprint arXiv:2008.05402},
  year={2020}
}

@inproceedings{SIDdataset,
  title={Sid: Stereo image dataset for autonomous driving in adverse conditions},
  author={El-Shair, Zaid A and Abu-raddaha, Abdalmalek and Cofield, Aaron and Alawneh, Hisham and Aladem, Mohamed and Hamzeh, Yazan and Rawashdeh, Samir A},
  booktitle={NAECON 2024-IEEE National Aerospace and Electronics Conference},
  pages={403--408},
  year={2024},
  organization={IEEE}
}

@article{TWICEdataset,
  title={Twice dataset: Digital twin of test scenarios in a controlled environment},
  author={Neto, Leonardo Novicki and Reway, Fabio and Poledna, Yuri and Drechsler, Maikol Funk and Icking, Christian and Huber, Werner and Ribeiro, Eduardo Parente},
  journal={International Journal of Vehicle Systems Modelling and Testing},
  volume={19},
  number={2},
  pages={152--170},
  year={2025},
  publisher={Inderscience Publishers (IEL)}
}

@article{WADSdataset,
  title={Winter adverse driving dataset for autonomy in inclement winter weather},
  author={Kurup, Akhil M and Bos, Jeremy P},
  journal={Optical Engineering},
  volume={62},
  number={3},
  pages={031207--031207},
  year={2023},
  publisher={Society of Photo-Optical Instrumentation Engineers}
}

@article{Yeong2021,
  title={Sensor and sensor fusion technology in autonomous vehicles: A review},
  author={Yeong, De Jong and Velasco-Hernandez, Gustavo and Barry, John and Walsh, Joseph},
  journal={Sensors},
  volume={21},
  number={6},
  pages={2140},
  year={2021},
  publisher={MDPI}
}

@article{STARNet,
  title={Starnet: Sensor trustworthiness and anomaly recognition via approximated likelihood regret for robust edge autonomy},
  author={Darabi, Nastaran and Tayebati, Sina and Ravi, Sathya and Tulabandhula, Theja and Trivedi, Amit R and others},
  journal={arXiv preprint arXiv:2309.11006},
  year={2023}
}

@inproceedings{RGB_CameraFailures,
  title={On failures of RGB cameras and their effects in autonomous driving applications},
  author={Secci, Francesco and Ceccarelli, Andrea},
  booktitle={2020 IEEE 31st International Symposium on Software Reliability Engineering (ISSRE)},
  pages={13--24},
  year={2020},
  organization={IEEE}
}

@inproceedings{OdometryRain,
  title={Evaluating visual odometry methods for autonomous driving in rain},
  author={Tan, Yu Xiang and Prasetyo, Marcel Bartholomeus and Daffa, Mohammad Alif and Nitin, Deshpande Sunny and Meghjani, Malika},
  booktitle={2023 IEEE 19th International Conference on Automation Science and Engineering (CASE)},
  pages={1--8},
  year={2023},
  organization={IEEE}
}

@article{MotionBlurRecovery,
  title={Towards robust visual odometry by motion blur recovery},
  author={Luan, Simin and Yang, Cong and Qin, Xue and Chen, Dongfeng and Sui, Wei},
  journal={Frontiers in Signal Processing},
  volume={4},
  pages={1417363},
  year={2024},
  publisher={Frontiers Media SA}
}

@article{Rob_LidarBased,
  title={Robustness of lidar-based pose estimation: Evaluating and improving odometry and localization under common point cloud corruptions},
  author={Yang, Bo and Triet Pham, Tri Minh and Yang, Jinqiu},
  journal={arXiv e-prints},
  pages={arXiv--2409},
  year={2024}
}

@inproceedings{Dong2023,
  title={Benchmarking robustness of 3D object detection to common corruptions in autonomous driving},
  author={Dong, Yinpeng and Kang, Caixin and Zhang, Jinlai and Zhu, Zijian and Wang, Yikai and Yang, Xiao and Su, Hang and Wei, Xingxing and Zhu, Jun},
  booktitle={2023 IEEE/CVF Conference on Computer Vision and Pattern Recognition (CVPR)},
  pages={1022--1032},
  year={2023},
  organization={IEEE}
}

@inproceedings{Chang2023,
  title={Failure mode investigation to enable lidar health monitoring for automotive application},
  author={Chang, Fred and Jafarzadeh, Ehsan and Del Gatto, Jacqueline and Cran, Graham and Sadjadi, Hossein},
  booktitle={Annual Conference of the PHM Society},
  volume={15},
  number={1},
  year={2023}
}

@article{Dust_Radar,
  title={Effects of dust and moisture surface contaminants on automotive radar sensor frequencies},
  author={Kang, Jeongmin and Hamidi, Oskar and Van{\"a}s, Karl and Eidev{\aa}g, Tobias and Nilsson, Emil and Friel, Ross},
  journal={Sensors},
  volume={25},
  number={7},
  pages={2192},
  year={2025},
  publisher={MDPI}
}

@inproceedings{Radar_Artifacts,
  title={Modeling and simulation of radar sensor artifacts for virtual testing of autonomous driving},
  author={Holder, Martin F and Linnhoff, Clemens and Rosenberger, Philipp and Popp, Christoph and Winner, Hermann},
  booktitle={9. Tagung Automatisiertes Fahren},
  year={2019}
}

@inproceedings{Radar_anomaly,
  title={Anomaly detection in radar data using PointNets},
  author={Griebel, Thomas and Authaler, Dominik and Horn, Markus and Henning, Matti and Buchholz, Michael and Dietmayer, Klaus},
  booktitle={2021 IEEE International Intelligent Transportation Systems Conference (ITSC)},
  pages={2667--2673},
  year={2021},
  organization={IEEE}
}

@inproceedings{FaultResistantOdometry,
  title={Fault resistant odometry estimation using message passing neural network},
  author={Dahal, Pragyan and Mentasti, Simone and Paparusso, Luca and Arrigoni, Stefano and Braghin, Francesco},
  booktitle={2023 IEEE Intelligent Vehicles Symposium (IV)},
  pages={1--8},
  year={2023},
  organization={IEEE}
}

@inproceedings{RedundantOdometry,
  title={Simple but effective redundant odometry for autonomous vehicles},
  author={Reinke, Andrzej and Chen, Xieyuanli and Stachniss, Cyrill},
  booktitle={2021 IEEE International Conference on Robotics and Automation (ICRA)},
  pages={9631--9637},
  year={2021},
  organization={IEEE}
}

@inproceedings{A2DO,
  title={A2DO: adaptive anti-degradation odometry with deep Multi-Sensor fusion for autonomous navigation},
  author={Lai, Hui and Chen, Qi and Zhang, Junping and Pu, Jian},
  booktitle={2025 IEEE International Conference on Robotics and Automation (ICRA)},
  pages={8589--8595},
  year={2025},
  organization={IEEE}
}

@article{Sensor_fault_diagnosis,
  title={Deep learning based data fusion for sensor fault diagnosis and tolerance in autonomous vehicles},
  author={Pan, Huihui and Sun, Weichao and Sun, Qiming and Gao, Huijun},
  journal={Chinese journal of mechanical engineering},
  volume={34},
  number={1},
  pages={72},
  year={2021},
  publisher={Springer}
}

@article{TEFuNet,
  title={TEFu-Net: A time-aware late fusion architecture for robust multi-modal ego-motion estimation},
  author={Agostinho, Lucas and Pereira, Diogo and Hiolle, Antoine and Pinto, Andry},
  journal={Robotics and Autonomous Systems},
  volume={177},
  pages={104700},
  year={2024},
  publisher={Elsevier}
}

@inproceedings{TransFusion,
  title={Transfusion: Robust lidar-camera fusion for 3d object detection with transformers},
  author={Bai, Xuyang and Hu, Zeyu and Zhu, Xinge and Huang, Qingqiu and Chen, Yilun and Fu, Hongbo and Tai, Chiew-Lan},
  booktitle={Proceedings of the IEEE/CVF conference on computer vision and pattern recognition},
  pages={1090--1099},
  year={2022}
}

@inproceedings{MultiModalExpert,
  title={Resilient sensor fusion under adverse sensor failures via multi-modal expert fusion},
  author={Park, Konyul and Kim, Yecheol and Kim, Daehun and Choi, Jun Won},
  booktitle={2025 IEEE/CVF Conference on Computer Vision and Pattern Recognition (CVPR)},
  pages={6720--6729},
  year={2025},
  organization={IEEE}
}

@article{Robust_localization_weather,
  title={Deep learning-based robust positioning for all-weather autonomous driving},
  author={Almalioglu, Yasin and Turan, Mehmet and Trigoni, Niki and Markham, Andrew},
  journal={Nature machine intelligence},
  volume={4},
  number={9},
  pages={749--760},
  year={2022},
  publisher={Nature Publishing Group UK, London}
}

@article{BEVFusion,
  title={Bevfusion: A simple and robust lidar-camera fusion framework},
  author={Liang, Tingting and Xie, Hongwei and Yu, Kaicheng and Xia, Zhongyu and Lin, Zhiwei and Wang, Yongtao and Tang, Tao and Wang, Bing and Tang, Zhi},
  journal={Advances in neural information processing systems},
  volume={35},
  pages={10421--10434},
  year={2022}
}

@article{LidarGenerico,
  title={Lidar for autonomous driving: The principles, challenges, and trends for automotive lidar and perception systems},
  author={Li, You and Ibanez-Guzman, Javier},
  journal={IEEE Signal Processing Magazine},
  volume={37},
  number={4},
  pages={50--61},
  year={2020},
  publisher={IEEE}
}

@article{AdverseWeatherSurvey,
  title={Perception and sensing for autonomous vehicles under adverse weather conditions: A survey},
  author={Zhang, Yuxiao and Carballo, Alexander and Yang, Hanting and Takeda, Kazuya},
  journal={ISPRS Journal of Photogrammetry and Remote Sensing},
  volume={196},
  pages={146--177},
  year={2023},
  publisher={Elsevier}
}

@article{Srivastav2023,
  title={Radars for autonomous driving: A review of deep learning methods and challenges},
  author={Srivastav, Arvind and Mandal, Soumyajit},
  journal={IEEE Access},
  volume={11},
  pages={97147--97168},
  year={2023},
  publisher={IEEE}
}

@article{Giuffrida2023,
  title={A survey of automotive radar and lidar signal processing and architectures},
  author={Giuffrida, Luigi and Masera, Guido and Martina, Maurizio},
  journal={Chips},
  volume={2},
  number={4},
  pages={243--261},
  year={2023},
  publisher={MDPI}
}

@inproceedings{Rao2025,
  title={Signal processing challenges in automotive radar},
  author={Rao, Sandeep and Narasimha, Rajan and Sun, Shunqiao},
  booktitle={ICASSP 2025-2025 IEEE International Conference on Acoustics, Speech and Signal Processing (ICASSP)},
  pages={1--5},
  year={2025},
  organization={IEEE}
}

@article{Baseline,
  title={A hybrid fault diagnosis method for autonomous driving sensing systems based on information complexity},
  author={Jin, Tianshi and Zhang, Chenxi and Zhang, Yikang and Yang, Mingliang and Ding, Weiping},
  journal={Electronics},
  volume={13},
  number={2},
  pages={354},
  year={2024},
  publisher={MDPI}
}

@article{MultiSensorFaultDetection,
  title={Multi-sensor fault detection, identification, isolation and health forecasting for autonomous vehicles},
  author={Safavi, Saeid and Safavi, Mohammad Amin and Hamid, Hossein and Fallah, Saber},
  journal={Sensors},
  volume={21},
  number={7},
  pages={2547},
  year={2021},
  publisher={MDPI}
}

@book{CameraComponents,
  title={Camera image quality benchmarking},
  author={Phillips, Jonathan B and Eliasson, Henrik},
  year={2018},
  publisher={John Wiley \& Sons}
}

@article{ISP_meets_DeepLearning,
  title={ISP meets deep learning: A survey on deep learning methods for image signal processing},
  author={da Silva, Matheus Henrique Marques and da Silva, Jhessica Victoria Santos and Arrais, Rodrigo Reis and Neto, Wladimir Barroso Guedes de Ara{\~A}{\v{s}}jo and Lopes, Leonardo Tadeu and Bileki, Guilherme Augusto and Lima, Iago Oliveira and Rondon, Lucas Borges and de Souza, Bruno Melo and Regazio, Mayara Costa and others},
  journal={arXiv preprint arXiv:2305.11994},
  year={2023}
}

@inproceedings{DynamicISP,
  title={Dynamicisp: dynamically controlled image signal processor for image recognition},
  author={Yoshimura, Masakazu and Otsuka, Junji and Irie, Atsushi and Ohashi, Takeshi},
  booktitle={2023 IEEE/CVF International Conference on Computer Vision (ICCV)},
  pages={12820--12830},
  year={2023},
  organization={IEEE}
}

@article{Radar_Sediments,
  title={Signal degradation through sediments on safety-critical radar sensors},
  author={Ehrnsperger, Matthias G and Siart, Uwe and Moosb{\"u}hler, Michael and Daporta, Emil and Eibert, Thomas F},
  journal={Advances in Radio Science},
  volume={17},
  pages={91--100},
  year={2019},
  publisher={Copernicus Publications, G{\"o}ttingen, Germany}
}

@article{RadarGracefulDegradation,
  title={Radar detection performability under graceful degradation},
  author={Ridder, Tyler D and Narayanan, Ram M},
  journal={International Journal of Performability Engineering},
  volume={17},
  number={8},
  pages={666},
  year={2021}
}

@article{RadarMisalignment,
  title={Overview of radar alignment methods and analysis of radar misalignment’s impact on active safety and autonomous systems},
  author={Burza, Rafa{\l} Micha{\l}},
  journal={Sensors},
  volume={24},
  number={15},
  pages={4913},
  year={2024},
  publisher={MDPI}
}

@article{Thermal_man_Radar,
  title={Thermal management of automotive radar: Overcoming design challenges in constrained environments},
  author={Oh, Myong Hun and Yoon, Su Hyun and Jeon, Soo Il and Choi, Minsuk},
  journal={Microelectronics Reliability},
  volume={168},
  pages={115725},
  year={2025},
  publisher={Elsevier}
}

@inproceedings{kong2023robo3d,
    author    = {Lingdong Kong and Youquan Liu and Xin Li and Runnan Chen and Wenwei Zhang and Jiawei Ren and Liang Pan and Kai Chen and Ziwei Liu},
    title     = {{Robo3D}: Towards Robust and Reliable {3D} Perception against Corruptions},
    booktitle = {Proceedings of the IEEE/CVF International Conference on Computer Vision (ICCV)},
    pages     = {19994-20006},
    year      = {2023}
}

@inproceedings{sensorsFog,
  title={Seeing through fog without seeing fog: Deep multimodal sensor fusion in unseen adverse weather},
  author={Bijelic, Mario and Gruber, Tobias and Mannan, Fahim and Kraus, Florian and Ritter, Werner and Dietmayer, Klaus and Heide, Felix},
  booktitle={2020 IEEE/CVF Conference on Computer Vision and Pattern Recognition (CVPR)},
  pages={11679--11689},
  year={2020},
  organization={IEEE}
}

@inproceedings{LiDAR_fog,
  title={A benchmark for lidar sensors in fog: Is detection breaking down?},
  author={Bijelic, Mario and Gruber, Tobias and Ritter, Werner},
  booktitle={2018 IEEE intelligent vehicles symposium (IV)},
  pages={760--767},
  year={2018},
  organization={IEEE}
}

@article{LiDAR_snow,
  title={4denoisenet: Adverse weather denoising from adjacent point clouds},
  author={Sepp{\"a}nen, Alvari and Ojala, Risto and Tammi, Kari},
  journal={IEEE Robotics and Automation Letters},
  volume={8},
  number={1},
  pages={456--463},
  year={2022},
  publisher={IEEE}
}

@article{LiDAR_Dust,
  title={When the dust settles: The four behaviors of lidar in the presence of fine airborne particulates},
  author={Phillips, Tyson Govan and Guenther, Nicky and McAree, Peter Ross},
  journal={Journal of field robotics},
  volume={34},
  number={5},
  pages={985--1009},
  year={2017},
  publisher={Wiley Online Library}
}

@inproceedings{LiDAR_Crosstalk,
  title={A methodology for analyzing the impact of crosstalk on lidar measurements},
  author={Brinon-Arranz, Lara and Rakotovao, Tiana and Creuzet, Thierry and Karaoguz, Cem and El-Hamzaoui, Oussama},
  booktitle={2021 IEEE Sensors},
  pages={1--4},
  year={2021},
  organization={IEEE}
}

@article{LiDAR_Wet,
  title={Characteristics of laser backscattering intensity to detect frozen and wet surfaces on roads},
  author={Shin, Jungil and Park, Hyunsuk and Kim, Taejung},
  journal={Journal of Sensors},
  volume={2019},
  number={1},
  pages={8973248},
  year={2019},
  publisher={Wiley Online Library}
}

@book{OutdoorCorrosion,
  editor    = {H. E. Townsend},
  title     = {Outdoor Atmospheric Corrosion},
  publisher = {ASTM International},
  address   = {West Conshohocken, PA, USA},
  year      = {2002}
}

@inproceedings{ChromaticAberration,
  title={Automatic removal of chromatic aberration from a single image},
  author={Kang, Sing Bing},
  booktitle={2007 IEEE Conference on Computer Vision and Pattern Recognition},
  pages={1--8},
  year={2007},
  organization={IEEE}
}

@inproceedings{Matos2025SimulatingSensorFailures,
  title={Simulating the effects of sensor failures on autonomous vehicles for safety evaluation},
  author={Matos, Francisco and Dur{\~a}es, Jo{\~a}o and Cunha, Jo{\~a}o},
  booktitle={Informatics},
  volume={12},
  number={3},
  pages={94},
  year={2025},
  organization={MDPI}
}

@article{Radar_software,
  title={Radar-stda: A high-performance spatial-temporal denoising autoencoder for interference mitigation of fmcw radars},
  author={Liu, Lulu and Guan, Runwei and Ma, Fei and Smith, Jeremy and Yue, Yutao},
  journal={arXiv preprint arXiv:2307.09063},
  year={2023}
}

@inproceedings{Radar_software2,
  author    = {Max Sponner and Julius Ott and Lorenzo Servadei and Bernd Waschneck and Robert Wille and Akash Kumar},
  title     = {Temporal Patience: Efficient Adaptive Deep Learning for Embedded Radar Data Processing},
  booktitle = {Proceedings of the IEEE International Conference on Embedded and Ubiquitous Computing (EUC)},
  year      = {2023},
  publisher = {IEEE},
  doi       = {10.1109/EUC58882.2023.00045}
}

@article{matos2024survey,
  title={A survey on sensor failures in autonomous vehicles: Challenges and solutions},
  author={Matos, Francisco and Bernardino, Jorge and Dur{\~a}es, Jo{\~a}o and Cunha, Jo{\~a}o},
  journal={Sensors},
  volume={24},
  number={16},
  pages={5108},
  year={2024},
  publisher={MDPI}
}

@article{viktor2026modality,
  title={From Modality Performance to Graceful Degradation: A PRISMA 2020 Systematic Review of Sensor Architectures for Autonomous Vehicles},
  author={Viktor, Patrik},
  journal={Sensors},
  volume={26},
  number={16},
  pages={5316},
  year={2026},
  publisher={MDPI}
}

@inproceedings{chen2019selective,
  title={Selective sensor fusion for neural visual-inertial odometry},
  author={Chen, Changhao and Rosa, Stefano and Miao, Yishu and Lu, Chris Xiaoxuan and Wu, Wei and Markham, Andrew and Trigoni, Niki},
  booktitle={2019 IEEE/CVF conference on computer vision and pattern recognition (CVPR)},
  pages={10534--10543},
  year={2019},
  organization={IEEE}
}

@article{chen2022learning,
  title={Learning selective sensor fusion for state estimation},
  author={Chen, Changhao and Rosa, Stefano and Lu, Chris Xiaoxuan and Wang, Bing and Trigoni, Niki and Markham, Andrew},
  journal={IEEE Transactions on Neural Networks and Learning Systems},
  volume={36},
  number={3},
  pages={4103--4117},
  year={2022},
  publisher={IEEE}
}

@article{talebi2025raci,
  title={Raci-Net: Ego-vehicle Odometry Estimation in Adverse Weather Conditions},
  author={Talebi, Mohammadhossein and Dahal, Pragyan and Possenti, Davide and Arrigoni, Stefano and Braghin, Francesco},
  journal={arXiv preprint arXiv:2507.10376},
  year={2025}
}

@standard{ISO26262_1,
  title        = {Road vehicles --- Functional safety --- Part 1: Vocabulary},
  organization = {International Organization for Standardization},
  number       = {ISO 26262-1:2018},
  year         = {2018}
}

@techreport{ITU838,
  author       = {{International Telecommunication Union, Radiocommunication Sector}},
  title        = {Recommendation {ITU-R} {P.838-3}: Specific attenuation model for rain for use in prediction methods},
  institution  = {International Telecommunication Union},
  address      = {Geneva, Switzerland},
  year         = {2005},
  month        = mar,
  url          = {https://www.itu.int/rec/R-REC-P.838-3-200503-I/en}
}

@article{lee2016visibility,
  title={Visibility: how applicable is the century-old Koschmieder model?},
  author={Lee, Zhongping and Shang, Shaoling},
  journal={Journal of the Atmospheric Sciences},
  volume={73},
  number={11},
  pages={4573--4581},
  year={2016}
}

@article{horvath1971applicability,
  title={On the applicability of the Koschmieder visibility formula},
  author={Horvath, Helmuth},
  journal={Atmospheric Environment (1967)},
  volume={5},
  number={3},
  pages={177--184},
  year={1971},
  publisher={Elsevier}
}

@inproceedings{jha2018avfi,
  title={Avfi: Fault injection for autonomous vehicles},
  author={Jha, Saurabh and Banerjee, Subho S and Cyriac, James and Kalbarczyk, Zbigniew T and Iyer, Ravishankar K},
  booktitle={2018 48th annual ieee/ifip international conference on dependable systems and networks workshops (dsn-w)},
  pages={55--56},
  year={2018},
  organization={IEEE}
}

@inproceedings{jha2019ml,
  title={Ml-based fault injection for autonomous vehicles: A case for bayesian fault injection},
  author={Jha, Saurabh and Banerjee, Subho and Tsai, Timothy and Hari, Siva KS and Sullivan, Michael B and Kalbarczyk, Zbigniew T and Keckler, Stephen W and Iyer, Ravishankar K},
  booktitle={2019 49th annual IEEE/IFIP international conference on dependable systems and networks (DSN)},
  pages={112--124},
  year={2019},
  organization={IEEE}
}

@inproceedings{maleki2023carfase,
  title={CarFASE: A carla-based tool for evaluating the effects of faults and attacks on autonomous driving stacks},
  author={Maleki, Mehdi and Farooqui, Ashfaq and Sangchoolie, Behrooz},
  booktitle={2023 53rd Annual IEEE/IFIP International Conference on Dependable Systems and Networks Workshops (DSN-W)},
  pages={92--99},
  year={2023},
  organization={IEEE}
}

@techreport{ITU833,
  author      = {{International Telecommunication Union, Radiocommunication Sector}},
  title       = {Recommendation {ITU-R} {P.833-10}: Attenuation in vegetation},
  institution = {International Telecommunication Union},
  address     = {Geneva, Switzerland},
  year        = {2021}
}

@article{Kettelgerdes2026,
  title={Precise intrinsic calibration and geometrical stability analysis of automotive solid-state LiDAR sensors over temperature and lifetime},
  author={Kettelgerdes, Marcel and Louwe Kooijmans, Evgeniia and Erdogan, H{\"u}seyin and Wunderle, Bernhard and Elger, Gordon},
  journal={Optics and Lasers in Engineering},
  volume={202},
  pages={109761},
  year={2026},
  publisher={Elsevier},
  doi={10.1016/j.optlaseng.2026.109761}
}

@article{HarterRadar,
  title={Self-Calibration of a 3-D-Digital Beamforming Radar System for Automotive Applications with Installation Behind Automotive Covers},
  author={Harter, Marlene and Hildebrandt, J{\"u}rgen and Ziroff, Andreas and Zwick, Thomas},
  journal={IEEE Transactions on Microwave Theory and Techniques},
  volume={64},
  number={9},
  pages={2994--3000},
  year={2016},
  publisher={IEEE},
  doi={10.1109/TMTT.2016.2593731}
}

@article{Albrecht2022,
  title={Warpage measurements to support the development of mmWave modules},
  author={Albrecht, Oliver and Meier, Karsten and Schaulin, Michael and Toussaint, Pierre-Louis and Gamba, Baptiste and Cruz, Rodolfo and Vernhes, Pierre and G{\"o}tze, Christian and Bock, Karlheinz},
  journal={Microelectronics Reliability},
  volume={138},
  pages={114731},
  year={2022},
  publisher={Elsevier},
  doi={10.1016/j.microrel.2022.114731}
}
